%% file: main.tex
\documentclass[lettersize,journal]{IEEEtran}

\usepackage{times}
\usepackage{epsfig}
\usepackage{graphicx}
\usepackage{svg}
\usepackage{amsmath}
\usepackage{amssymb}
\usepackage[noend]{algpseudocode}
\usepackage{balance}
\usepackage{url}
\usepackage{authblk}
\usepackage{multirow}
\usepackage{hyperref}
\hypersetup{
    colorlinks=true,
    linkcolor=blue,
    filecolor=magenta,      
    urlcolor=cyan,
    pdftitle={},
    pdfpagemode=FullScreen,
    }

\newcommand{\ie}{\emph{i.e.},}
\newcommand{\etal}{\emph{et~al.}}

\def\eqref#1{Equation~(\ref{#1})}

\usepackage{cite}
\usepackage[font=small]{caption}
\usepackage{subcaption}
\usepackage{graphbox}
\usepackage{soul}

\usepackage{color, colortbl}
\usepackage{tabularx}

\usepackage[linesnumbered,ruled]{algorithm2e}
\usepackage[flushleft]{threeparttable}
\usepackage{lipsum}
\usepackage{amssymb}%
\usepackage{pifont}%
\newcommand\numberthis[1][]{%
    \refstepcounter{equation}%
    \ifx#1\empty\else\label{eq:#1}\fi%
    \tag{\theequation}%
}

\usepackage{caption}

\usepackage{colortbl}
\usepackage{xcolor}

\makeatletter
\def\BState{\State\hskip-\ALG@thistlm}
\makeatother
\algdef{SE}[DOWHILE]{Do}{doWhile}{\algorithmicdo}[1]{\algorithmicwhile\ #1}

\graphicspath{ {figures/} }

\title{\LARGE \bf
{\textit{FactoFormer:} Factorized Hyperspectral Transformers with Self-Supervised Pretraining}
}

\author{Shaheer Mohamed$^{1,2}$,~\IEEEmembership{Student Member,~IEEE,} Maryam Haghighat$^{2,1}$,~\IEEEmembership{Member,~IEEE,} Tharindu Fernando$^{2}$,~\IEEEmembership{Member,~IEEE,} Sridha Sridharan$^{2}$,~\IEEEmembership{Life Senior Member,~IEEE,} 
Clinton Fookes$^{2}$,~\IEEEmembership{Senior Member,~IEEE}, Peyman Moghadam$^{1,2}$,~\IEEEmembership{Senior Member,~IEEE}   

\thanks{Manuscript accepted by IEEE Transactions on Geoscience and Remote Sensing in December 2023.}

\thanks{
$^1$ Shaheer Mohamed, Maryam Haghighat and Peyman Moghadam are with the Robotics and Autonomous Systems, DATA61, CSIRO, Brisbane, QLD 4069, Australia. 
E-mails: {\tt\footnotesize \emph{firstname.lastname}@data61.csiro.au} }
\thanks{
$^{2}$ Shaheer Mohamed, Maryam Haghighat, Tharindu Fernando, Sridha Sridharan,  Clinton Fookes, and Peyman Moghadam are with the research program in Signal Processing, Artificial Intelligence and Vision Technologies (SAIVT) at the Queensland University of Technology (QUT), Brisbane, Australia.
E-mails: {\tt\footnotesize \emph{shaheer.mohamed, maryam.haghighat, t.warnakulasuriya, s.sridharan, c.fookes, peyman.moghadam}@qut.edu.au}}
 }

\usepackage{arydshln}
\newcolumntype{x}[1]{>{\centering\arraybackslash\hspace{0pt}}p{#1}}
\newcolumntype{M}[1]{>{\centering\arraybackslash}m{#1}}
\newcolumntype{L}[1]{>{\raggedright\arraybackslash} m{#1} }

\usepackage{titlesec}
\titlespacing*{\section}{0pt}{4mm}{4mm}

\begin{document}

\markboth{IEEE TRANSACTIONS ON GEOSCIENCE AND REMOTE SENSING}%
{Shell \MakeLowercase{\textit{et al.}}: A Sample Article Using IEEEtran.cls for IEEE Journals}

\maketitle
\input{chapters/abstract_2.tex}


\input{chapters/introduction-new}

\input{chapters/relatedworks}

\input{chapters/background}

\input{chapters/methodology}

\input{chapters/experiments}

\input{chapters/results}

\input{chapters/discussion}

\input{chapters/conclusion.tex}

\section*{Acknowledgements}
The authors gratefully acknowledge funding of the project by the CSIRO and QUT. This work was partially funded by CSIRO's Digital Agriculture Scholarship and CSIRO's Machine Learning and Artificial Intelligence Future Science Platform (MLAI FSP).

\balance{}

{\small
        \bibliographystyle{IEEEtran}
        \bibliography{main}

}

\end{document}

%% file: chapters/abstract_2.tex
\begin{abstract}

Hyperspectral images (HSIs) contain rich spectral and spatial information. Motivated by the success of transformers in the field of natural language processing and computer vision where they have shown the ability to learn long range dependencies within input data, recent research has focused on using transformers for HSIs. However, current state-of-the-art hyperspectral transformers only tokenize the input HSI sample along the spectral dimension, resulting in the under-utilization of spatial information. Moreover, transformers are known to be data-hungry and their performance relies heavily on large-scale pretraining, which is challenging due to limited annotated hyperspectral data. Therefore, the full potential of HSI transformers has not been fully realized. To overcome these limitations, we propose a novel factorized spectral-spatial transformer that incorporates factorized self-supervised pretraining procedures, leading to significant improvements in performance. The factorization of the inputs allows the spectral and spatial transformers to better capture the interactions within the hyperspectral data cubes. Inspired by masked image modeling pretraining, we also devise efficient masking strategies for pretraining each of the spectral and spatial transformers. We conduct experiments on six publicly available datasets for HSI classification task and demonstrate that our model achieves state-of-the-art performance in all the datasets. The code for our model will be made available at \href{https://github.com/csiro-robotics/factoformer}{https://github.com/csiro-robotics/FactoFormer}.

\end{abstract}

%% file: chapters/introduction-new.tex
\section{Introduction}
\label{sec:intro}

\IEEEPARstart{H}{yperspectral} Images (HSI) capture data from a wide range of the electromagnetic spectrum at each pixel. Compared to RGB and multi-spectral images, HSIs contain more fine-grained spectral information, making them widely used in applications such as precision agriculture \cite{moghadam2017plant}, environment monitoring\cite{GUO2023120}, security \cite{military}, urban environment planning \cite{urban_planning}, geology \cite{geology}, food quality analysis \cite{food-quality}, and medical \cite{medical}. Hyperspectral cameras capture reflectance spectra at each pixel, resulting in three-dimensional data composed of spatial and spectral information. The spatial domain consists of information such as shape, texture, and object layout, while the spectral domain is composed of reflectance spectra with materialistic or physical properties. 

In the early days of deep learning, Convolution Neural Networks (CNNs) were widely used for hyperspectral images\cite{1D-CNN,2D-CNN, chen2016deep, roy2019hybridsn, mahendren2021reduction}. However, CNNs are not well-suited for capturing long-range dependencies within the spectral domain of HSI, which can span several hundred spectral bands.
Recently, Transformers have revolutionized the field of natural language processing and computer vision by effectively learning long-range dependencies within input data, and their success has extended to HSI as well ~\cite{hong2021spectralformer, MAEST, SSFT}.

However, previous methods ~\cite{hong2021spectralformer, MAEST} mainly applied pure Vision Transformer (ViT) architecture for spectral encoding, using a simple yet effective spectral group tokenization strategy, but they have not fully exploited the spatial and spectral interactions present in hyperspectral image cubes. 
Furthermore, the limitation of annotated data is a significant challenge when training ViTs, as they rely heavily on large-scale training data. However, most of the available hyperspectral datasets contain limited labeled proportions, such as the Indian Pines, University of Pavia, and Houston 2013 datasets, which possess only 48.75\%, 17.43\%, and 2.26\% of annotated data, respectively. Recently, Masked Image Modeling (MIM)~\cite{MAE, SimMIM} has been introduced as a simple yet powerful self-supervised pretraining framework for Vision Transformers (ViT) without the use of labels. However, the MIM pretraining strategies often have been limited to a single modality (RGB). 

To tackle the aforementioned limitations, we propose \textit{FactoFormer}: a factorized transformer architecture for hyperspectral images that leverages interactions along the spectral and spatial dimensions of hyperspectral cubes. \textit{FactoFormer} splits hyperspectral image cubes into non-overlapping tokenized patches along both spectral and spatial dimensions and processes them with two transformers that individually focus on spatial and spectral dimensions, namely the spectral transformer, and spatial transformer. The factorization enables factorized self-attention where attention is computed in spatial and spectral dimensions simultaneously to learn and extract more salient information in both dimensions. Finally, we fuse these learned latent representations from both spectral and spatial transformers to extract the intrinsic higher-order spectral-spatial correlations present in hyperspectral data. Since the computation of multi-head self-attention in transformers is associated with quadratic computational complexity, jointly processing them would not be ideal as it increases the computational overhead. In \textit{FactoFormer}, by factorization we not only learn the spectral and spatial features more effectively, but we also reduce the computational complexity from $O((m+n)^2)$ to $O(m)^2 + O(n)^2)$ where $n$ and $m$ are the number of spatial and spectral tokens respectively. The overall architecture of the proposed \textit{FactoFormer} is presented in Figure 1.

The proposed factorized architecture not only increases the efficiency and scalability of modeling spectral and spatial interactions but also enables factorized self-supervised pretraining. To mitigate the lack of large labeled data, we design novel masking strategies to effectively pretrain spectral and spatial transformers individually, leveraging a vast amount of unlabeled hyperspectral data. A random proportion of the input tokens are masked, and only the visible tokens are passed through spectral and spatial transformers, respectively. The latent space representations from each transformer, combined with the masked tokens, are then used to reconstruct the original information for the masked regions.

Experimental results on six large-scale hyperspectral datasets show that the proposed \textit{FactoFormer} achieves state-of-the-art performance while being computationally more efficient. Ablation studies were conducted to evaluate the impact of different components of our proposed method, providing insights into their contributions to the overall performance.

\begin{figure*}[t!]
    \centering
    \includegraphics[scale=0.85]{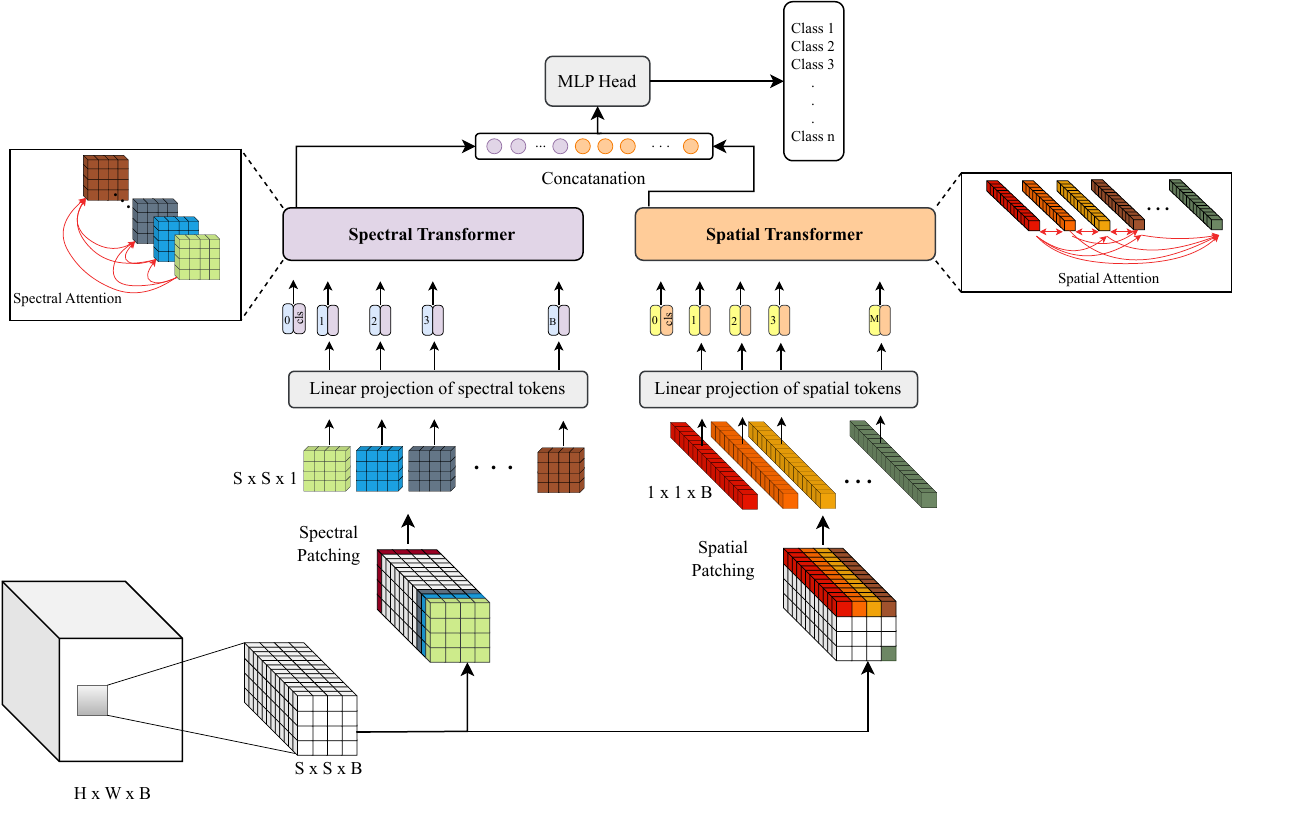}
    \caption{An overview of the proposed \textit{FactoFormer} architecture: a factorized transformer architecture for hyperspectral images. \textit{FactoFormer} splits hyperspectral image cubes into non-overlapping tokenized patches along spectral and spatial dimensions and processes them with two transformers simultaneously, where attention in each transformer focuses on spectral and spatial dimensions. The outputs of each transformer get combined by concatenating them and passing them to a multi-layer perceptron to perform classification.}
    \label{fig:overview}
\end{figure*}

The key contributions of the paper are summarized as follows:

\begin{itemize}
        \item We propose a factorized transformer architecture, called \textit{FactoFormer}, that leverages interactions along both spectral and spatial dimensions of hyperspectral cubes. 
        \item The proposed factorized architecture enables factorized self-attention that allows us to learn more salient features focusing on each dimension while reducing the computational complexity.
        \item We propose novel self-supervised pretraining strategies for the factorized spatial and spectral transformers, which allows us to better utilize the large proportion of unlabeled data in the public datasets.
        \item We evaluate our proposed \textit{FactoFormer} method using six publicly available benchmarks for HSI classification and our method outperforms state-of-the-art supervised and self-supervised transformer-based methods on all the datasets with significant margins.    
\end{itemize}

To showcase the effectiveness of the proposed \textit{FactoFormer} architecture, we conduct experiments on hyperspectral image classification using remote sensing data.  However, it is important to note that the \textit{FactoFormer} framework is not limited to HSI classification tasks and can be applied to various other HSI applications that require representation learning from both spatial and spectral information, even with limited annotated data. 
The remaining sections of the paper are organized as follows. Section~\ref{sec:relwork} provides a summary of related works and their limitations. Section~\ref{sec:background} outlines the background information. The proposed methodology is discussed in detail in Section~\ref{sec:method}. Section~\ref{sec:exp} outlines the experiments and implementation details. The results and ablation studies are presented in Section~\ref{sec:results}, and a discussion on merits and limitations of proposed method are included in Section \ref{sec:Discussion}. Finally, Section~\ref{sec:conclusion} concludes the work and discusses possible future directions.

%% file: chapters/relatedworks.tex
\section{Related Work}
\label{sec:relwork}
In this section, first, we discuss the overview of deep learning-based approaches for HSI classification. Following that, we highlight the limitations of CNN-based networks and present the motivation behind utilizing transformers for hyperspectral images. The section then delves into recent transformer-based networks used for HSI classification and their limitations. Finally, we explore the potential of self-supervised learning and how it can be leveraged to effectively address the scarcity of labeled data to train transformers.

\subsection{Deep Learning Based Approaches}

As deep learning models learn better feature representations from the data itself without using any prior knowledge, they result in state-of-the-art performance. Extensive research has been conducted in the past decade with various deep learning-based network architectures for better feature extraction in hyperspectral data. Chen \etal{}~\cite{AE-chen2014deep} proposed a Stacked-Auto-Encoder (SAE) based network for HSI classification. Their proposed model is a combination of principal component analysis (PCA), AE, and logistic regression. Recurrent Neural Networks (RNNs) are used for learning patterns in sequential data and
mostly used in language processing. Since spectral information in hyperspectral data contains sequential properties, Mou \etal{}~\cite{mou2017deep} proposed a modified gated recurrent unit (GRU) based framework that considers each pixel with its spectra as the input sequence to perform classification. Zohu \etal{}~\cite{zhou2019hyperspectral} considered both spatial and spectral features as sequences and proposed a long short-term memory (LSTM) based network for HSI classification. Most of the early deep learning based approaches are fully supervised and depends on labelled data. A major concern when applying deep learning based solutions to HSIs is the limitation of annotated data. To address this problem, different techniques like self-supervised learning \cite{MAEST, MAE, DINO} and domain adaptation \cite{ref_2, ref_3, ref_5} have been explored in recent literature.   

\subsection{CNN-based Networks for Hyperspectral Images}
Convolutional neural networks (CNNs) are the most prominent network architecture in remote sensing for HSI classification. Many works designed CNN-based networks for HSI classification using 1-D and 2-D CNNs \cite{1D-CNN,2D-CNN, ref_2}. Yang \etal{}~\cite{yang2017learning} proposed a dual-branch CNN network for extracting spectral and spatial branches separately and fusing the representations to perform classification. A 3D-CNN architecture was proposed by Chen \etal{}~\cite{chen2016deep}, where 3D convolutions are used to extract spatial-spectral features. Roy \etal{}~\cite{roy2019hybridsn} proposed a hybrid spectral CNN for hyperspectral classification where a combination of 3D and 2D CNNs are used for feature extraction, which reduced the computational complexity compared to 3D CNN-based models. Residual connections were introduced to CNN-based models to address the vanishing gradient problem with deep networks. Zhu \etal{}~\cite{9103247} proposed a CNN network with residual connections and soft attention to extracting better spectral-spatial features. Even though CNN-based networks have been successful in extracting contextual features, the inherent architecture fails to capture long-range dependencies. As hyperspectral data contains hundreds of bands with fine-grained information, it becomes crucial to capture these long-range dependencies to extract more distinctive features. Transformers have proven their effectiveness in both natural language processing and computer vision tasks, showcasing their ability to model long-range dependencies and extract meaningful semantic features. Moreover, hyperspectral data often exhibit substantial redundancy, and the multi-head self-attention mechanism within transformers computes pairwise interactions and emphasizes salient features while suppressing redundancy. Therefore, leveraging transformers for hyperspectral data holds significant potential for learning more meaningful features.

\subsection{Transformers for Hyperspectral Images}

Transformer-based networks have revolutionized the entire natural language processing domain with a significant performance boost. Inspired by this, vision transformer (ViT) \cite{vit_original} was introduced later, where images are treated as a sequence of patches. This model achieved state-of-the-art performance for image classification with large-scale pretraining. Transformers have proven to be the ideal architecture for capturing long range dependencies through its multi-head self attention mechanism which processes input as a sequence of tokens. Considering the large number of spectral bands in HSI, which has high correlation and granular information, extracting these global dependencies are crucial for learning better features. Therefore, utilizing transformers and adapting it for learning spatial-spectral features in HSI would lead to better performance. However, there are only few works exploring the transformer architectures for HSI and its full potential in the domain is yet to be explored. Sun \etal{}~\cite{SSFT} proposed a hybrid network that utilizes both CNNs and transformers. Initially, 3D and 2D CNNs are used to extract features, and then the extracted features are tokenized by Gaussian weighted tokenizer and sent to the transformer encoder. In this method, the main feature extraction is based on CNNs, where a single-layer transformer is utilized at the last layer to learn relationships between high-level semantic features. Mei \etal{}~\cite{GAHT} proposed a group-aware hierarchical transformer, which has three stages where each stage is composed of grouped pixel embedding and transformer layers. Here, the proposed grouped pixel embedding comprises convolutions to learn local features, while the transformer is utilized to learn global dependencies. These hybrid approaches perform the initial feature extraction using convolutions where the transformers are used in order to find the global dependency of extracted local features. They did not explore how transformers can be directly used with HSI to learn spectral-spatial feature interactions without convolutions. SpectralFormer \cite{hong2021spectralformer} utilized the ViT architecture and proposed a pure transformer-based network for HSI with an additional skip connection mechanism to fuse information between transformer layers.
Here, the input is patched along the spectra by grouping neighboring bands. Even though the model performed well, the input samples are patched only along the spectral dimension, where the spatial information in HSIs is not fully exploited. Also, the model is trained in a fully supervised manner using limited training samples, which degrades the performance of transformers due to the lack of annotated data. 
Recently, MAEST \cite{MAEST} utilizes a similar transformer architecture as SpectralFormer\cite{hong2021spectralformer} and added a self-supervised pretraining framework for only spectral dimension, which is briefly discussed in the next section. Peng \etal{}~\cite{CASST} proposed a spatial-spectral transformer architecture with cross-attention for HSI. Here, the input is tokenized in spatial and spectral dimensions and fused with a spatial-spectral cross-attention layer in the transformer block. In this method, principal component analysis is used initially to reduce the dimensionality; hence the model is not exposed to full spectral ranges which may cause the loss of fine-grained information. Also, the approach is fully supervised and trained with limited labeled data and the additional convolution layers used prior to tokenization adds more complexity to the model.

\subsection{Self-Supervised Learning}
Self-supervised learning (SSL) allows to learn good representations without large annotated databases \cite{DINO, BYOL, LiuSSL}. So far, most advances in deep learning have been dependent on supervised learning methods where high-quality labeled data is available. However, in many applications, such as hyperspectral image classification, there are limited annotated data sources. SSL is a self-learning mechanism that leverages the underlying structure in the data using labels that are generated from the data itself. Most of the previous methods in SSL are based on contrastive learning approaches in which representations are learned by bringing positive samples closer and pushing away the negative samples in embedding space \cite{SimCLR, DINO}. These methods depend on large memory banks and heavy augmentations. Masked Image Modeling (MIM) approaches captures meaningful representations by predicting the masked-out regions of images and have shown promising results for ViTs~\cite{MAE, SimMIM}. Since meaningful augmentations in spectra of HSIs are challenging and also large negative samples cannot be generated due to data limitation, MIM approaches are more suitable for hyperspectral data compared to contrastive learning approaches. MAEST \cite{MAEST} proposes an autoencoder architecture inspired by MAE \cite{MAE} for hyperspectral data. The input is patched along the spectra and randomly masked, where the model learns feature representations by reconstructing the spectra. Finally, the encoder is fine-tuned for performing classification. MAEST\cite{MAEST} results in better performance compared to fully supervised approaches, but it learns only the spectral information with the transformer network and uses a computationally heavy decoder for reconstruction.

\vspace{\baselineskip}
In this section we have explored and analyzed related works that closely align with our proposed method. We propose \textit{FactoFormer}: a factorized transformer with self-supervised pretraining to learn rich spectral-spatial feature representations from hyperspectral data. The spectral-spatial factorization is proposed in \cite{CASST} as well, but the approach is fully supervised. Also, the input is pre-processed with PCA in order to reduce the spectral dimension, potentially leading to a loss of fine-grained information. The input is further processed with additional convolutions prior to sending it to the transformer. In contrast, \textit{FactoFormer} adopts a different approach. We tokenize the raw input without any preprocessing and directly feed it to the transformers. This approach enables the model to learn more fine-grained spatial-spectral correlations present in the hyperspectral data, making it well-suited for the task at hand.In terms of self-supervised pretraining both MAEST \cite{MAEST} and proposed \textit{FactoFormer} utilized the unlabelled data to pretrain the network. In MAEST \cite{MAEST}, the input is patched only along the spectra by grouping neighbouring bands, whereas in \textit{FactoFormer} we propose a factorized pretraining approach which enables learning rich spectral and spatial features separately during pretraining and jointly fine-tuning to learn the higher-order spectral-spatial features. Also compared to MAEST \cite{MAEST} our proposed method is computationally more efficient due to the simple linear layer decoder used to reconstruct only masked tokens whereas in MAEST \cite{MAEST} a heavy transformer-based decoder is used to reconstruct the full input.

%% file: chapters/background.tex
\section{Background}
\label{sec:background}

In this section, we briefly describe the ViT\cite{vit_original}, which is a modified version of the transformer \cite{vaswani2017attention} architecture that processes images as a sequence of patches. A 2D image  $X\in \mathbb{R}^{H \times W \times C}$ is divided into $N$ non-overlapping patches of $P\times P$ and flattened into $x_{p} \in \mathbb{R}^{(P^{2}\times C)}$ where $H, W, C$ are the height, width, and number of channels respectively. Then, $N= HW/P^2$ number of patches are transformed into $N$ embedding vectors  $z_{p} \in \mathbb{R}^{D}$ of size $D$ through a linear projection $\mathbf{E} \in \mathbb{R}^{D \times(P^{2} \times C)}$. The embedding size $D$ is constant in all layers, and the newly formed vector sequence is called patch embeddings. A  learnable positional embedding, $e_{p}^{pos}\in \mathbb{R}^D$, is added to each patch to encode the positional information as in Eq.~\ref{eq:pos_embed}:
\begin{equation}
    \mathrm{z}_{p} =\mathbf{E}x_p + e_{p}^{pos}, \quad p=1, ..., N. \newline
    \label{eq:pos_embed}
\end{equation}

Also, a learnable classification token $z_{cls} \in \mathbb{R}^D$ is added to the first position of the sequence, \ie{}  $ \mathrm{{z}_0}$, and trained together. Therefore, the updated sequence sent into the first layer of the transformer, $ \mathrm{z}^{(0)}$,  can be defined as in Eq.~\ref{eq:seq}: 

\begin{equation}
    \mathrm{z}^{(0)} = [ \mathrm{z_{cls}}, \mathrm{z_1}, \mathrm{z_2}, ... , \mathrm{z}_N ].
    \label{eq:seq}
\end{equation}

The classification token from the final layer of the transformer encoder is used in the classification layer. The transformer architecture consists of $\textit{L}$ layers, and each layer $l$ is composed of Multi-head Attention \cite{vaswani2017attention}, layer normalization (LN), and multi-layer perceptron (MLP) block. Multi-head attention is the key component of the transformer architecture that allows the network to generate rich feature representations. Multiple self-attention layers are stacked together and integrated to form multi-head attention, and this allows it to attend to different parts of the sequence and highlight salient information. 

Formally, each layer $l$ contains multiple attention heads. The output of the previous layer $\mathrm{z}^{(l-1)}$ is projected to query, key, and value matrices using $W_{q}^{i}, W_{k}^{i}, W_{v}^{i}  \in \mathbb{R}^{D \times D/h} $ learnable weight matrices as shown in Eq.~\ref{eq:qkv}:

\begin{equation}
    Q_{i} = \mathrm{z}^{(l-1)}W_q^{i}, \quad K_{i} = \mathrm{z}^{(l-1)}W_k^{i}, \quad V_{i} = \mathrm{z}^{(l-1)}W_v^{i} .
    \label{eq:qkv}
\end{equation}

The dot product of query and key matrices are computed, divided by ${\sqrt{d_k}}$, and passed through a softmax function to determine the attention score matrix. Here $d_k$ represents the dimension of $K$, which is equal to $D/h$, where $h$ is the number of heads. Finally, the attention score matrix is multiplied with the value matrix $V$ as shown in Eq.~\ref{eq:attention}, to compute the attention at each head $head_i$.  

\begin{equation}
    head_i=\mathrm{Attention(Q_i, K_i,V_i)} = softmax(\frac{Q_iK_i^{T}}{\sqrt{d_k}})V.   
    \label{eq:attention}
\end{equation}

All heads are concatenated and projected with $W^0$ matrix to get the output of multi-head attention at each layer as shown in Eq.~\ref{eq:multihead}. Next, the output is sent through an MLP layer which consists of two fully connected layers with a Gaussian error linear unit (GELU) activation function. This completes the computation for one transformer block. 
\begin{equation}
    \mathrm{MultiHead}(Q,K,V) = \mathrm{Concat}(head_1, ..., head_h)W^0.
    \label{eq:multihead}
\end{equation}

 Finally, a linear classifier is used with the classification token in the final layer $\mathrm{z}_{cls}^L $, or with global average pooling of all embeddings. 

 \vspace{2mm}

%% file: chapters/methodology.tex
\section{Methodology}
\label{sec:method}

In this section, we first give an overview of our proposed method \textit{FactoFormer}. Next, we discuss the proposed novel factorized transformer architecture. Finally, we explain the self-supervised pretraining strategy used in our method. The overall architecture of the model is shown in Figure \ref{fig:overview}.    

\subsection{FactoFormer}
\label{sec:factoformer}
Our proposed \textit{FactoFormer} learns spectral and spatial representations from hyperspectral data using factorized transformers.
We process the entire input cube using both spatial and spectral transformers in order to capture the intrinsic spatial-spectral correlations. But the way we tokenize and compute self-attention to learn salient spatial and spectral information discriminates against one another. Further, with the fusion mechanism employed at the end, we capture more fine-grained higher-order spatial-spectral correlations. The factorized architecture enables factorized pretraining, where both transformers can be treated separately and pretrained using specific self-supervised pre-text tasks.

For each labeled cube, an extended spatial neighborhood of $S \times S$ is selected as our input sample, $X \in  \mathbb{R}^{S \times S \times B}$,  where  $B$ is the number of bands in the HSI.We patch the input sample in both spatial and spectral dimensions and process them with two individual transformers. 

\vspace{0.1cm}
\noindent\textbf{Spectral Transformer:} The input sample $X$ is divided into $B$ number of non-overlapping patches along the spectra where each patch, $a_i \in \mathbb{R}^{S\times S \times 1}$, corresponds to a particular band in the spectrum and contains all the spatial information of the input sample. Next, we send these patches to the spectral transformer. First, patches are flattened as, $a_i \in \mathbb{R}^{S^2}$, and embedded to a fixed dimension, $d_{spe}$, using a linear projection matrix, $\mathbf{E}_{spe} \in \mathbb{R}^{d_{spe} \times S^2}$. Then positional embeddings, $e_{spe}^{pos} \in \mathbb{R}^{d_{spe}}$, are added as shown in Eq. \ref{eq:spe-proj}, to retain the positional information.

\begin{equation}
    \mathrm{z}_{(spe, i)}=\mathbf{E}_{(spe, i)} a_i + e_{(spe,i)}^{pos} \quad i=1,...,B.
    \label{eq:spe-proj}
\end{equation}

 An additional classification token, $\mathrm{z}_{(cls,spe)}\in \mathbb{R}^{d_{spe}}$ is added to this sequence, which learns an overall representation of the entire sequence and is used later for classification. The sequence of tokens, $\mathrm{z}_{spe}^{(0)}$, is sent to the spectral transformer as shown in Eq. \ref{eq:spe_seq}. The spectral transformer attends to the spectra and extracts salient information by suppressing noise and redundancy.

\begin{equation}
    \mathrm{z}_{spe}^{(0)} = [ \mathrm{z}_{(cls,spe)}, \mathrm{z_1}, \mathrm{z_2}, ... , \mathrm{z}_B ].
    \label{eq:spe_seq}
\end{equation}

\vspace{0.1cm}
\noindent\textbf{Spatial Transformer:} The input sample $X$ is divided into $M=S^2$ number of non-overlapping patches. Each patch is defined as $b_i \in \mathbb{R}^{1 \times 1 \times B} $, which contains the complete spectral information corresponding to each pixel. Next, we send this sequence of patches to the spatial transformer. First, the patches are flattened as, $b_i \in \mathbb{R}^{B}$, and projected to an embedding size of $d_{spa}$ using a linear projection matrix $\mathbf{E}_{spa} \in \mathbb{R}^{d_{spa} \times B} $, followed by adding positional embeddings, $e_{spa}^{pos} \in \mathbb{R}^{d_{spa}}$, as in Eq.~\ref{eq:spa-proj}.  

\begin{equation}
    \mathrm{z}_{(spa,i)}=\mathbf{E}_{(spa,i)} b_i + e_{(spa,i)}^{pos}
    \quad i=1,...,M.
    \label{eq:spa-proj}
\end{equation}

Then as shown in Eq.~\ref{eq:spa_seq}, the classification token, $\mathrm{z}_{(cls,spa)}\in \mathbb{R}^{d_{spa}}$ is added to the sequence, $\mathrm{z}_{spa}^{(0)}$, and sent to the spatial transformer. The spatial transformer attends to the spatial information of the input and learns respective correlations. 

\begin{equation}
    \mathrm{z}_{spa}^{(0)} = [ \mathrm{z}_{(cls,spa)}, \mathrm{z_1}, \mathrm{z_2}, ... , \mathrm{z}_M ].
    \label{eq:spa_seq}
\end{equation}

\vspace{0.1cm}
\noindent\textbf{Fusion:} Finally, learned embeddings from the factorized transformers are fused in order to perform classification. Since the classification token learns to represent the entire input sequence, the classification token in each transformer captures an overall context in both spectral and spatial dimensions. Therefore, we perform a simple concatenation of classification tokens, $\mathrm{z}^{(L)}_{(cls,spe)}, \mathrm{z}^{(L)}_{(cls,spa)}$ from the last layer, $L$, of both transformers and pass it through a multi-layer perceptron (MLP) to perform classification. The fusion of features from both spectral and spatial transformers ensures that the model learns to capture the higher-order spectral-spatial correlations present in HSI data.

\begin{figure}[t]
    \centering
    \includegraphics[scale=0.45]{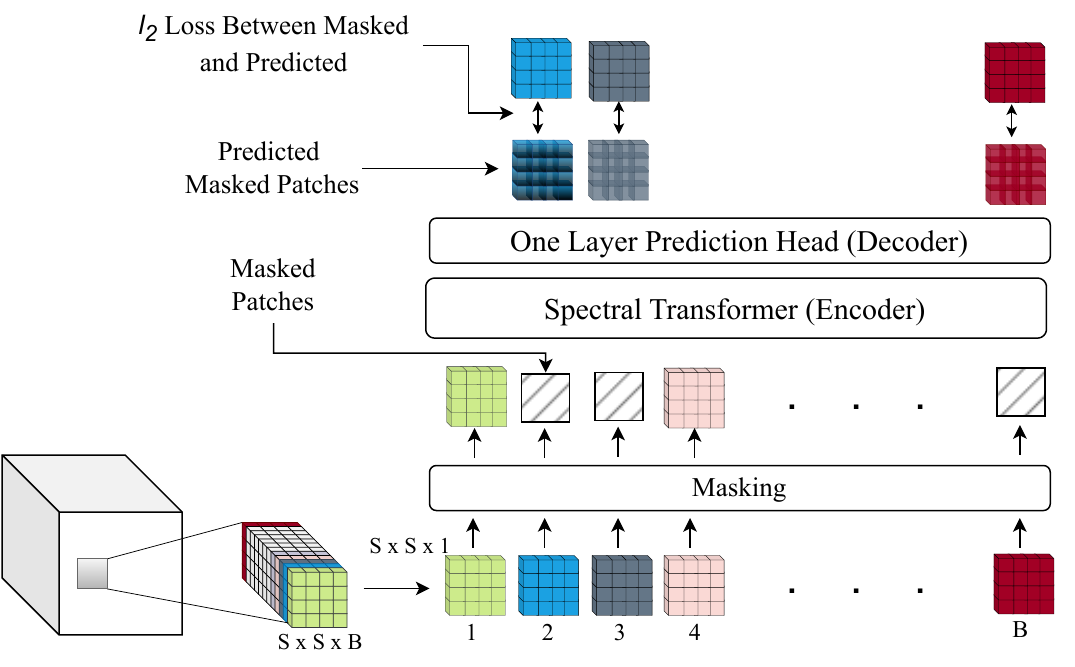}
    \caption{An overview of the proposed pretraining network for spectral transformer in \textit{FactoFormer}.}
    \label{fig:spectralSimMIM}
\end{figure}

\subsection{Self-Supervised Pretraining in FactoFormer}
Inspired by Masked Image Modeling (MIM), we design a new pretraining strategy for both spectral and spatial transformers. %
We utilized the large proportion of unlabeled data from the same dataset for self-supervised pretraining. As these unlabeled samples belong to the same image, environmental conditions, instrumental noise, and other affecting factors remain constant during pretraining. They are helpful for learning better feature representations. Both transformers are pretrained separately with spatially consistent and spectrally consistent masking strategies to enhance their underlying feature representations.

\vspace{0.1cm}
\noindent\textbf{Pretraining of Spectral Transformer:} We apply patching to the input sample in the spectral dimension, followed by linear embedding and positional encoding as described in Eq. \ref{eq:spe-proj}. Subsequently, we randomly mask tokens from the input sample, passing only the visible tokens through the spectral transformer. The latent space representations from the spectral transformer, combined with the masked tokens, are then used to reconstruct the original information for the masked regions. For this purpose, we utilize a simple linear layer as the decoder, in contrast to the heavy transformer-based decoder used in MAEST \cite{MAEST}. Finally, we use Mean Squared Error (MSE) as the loss function. 
As shown in Figure \ref{fig:spectralSimMIM}, masks are spatially consistent. Each token corresponds to the complete spatial information of a particular band in the input sample, allowing the model to learn better representations along the spectral dimension from spectrally patched tokens.

\begin{figure}[t]
    \centering
    \includegraphics[scale=0.42]{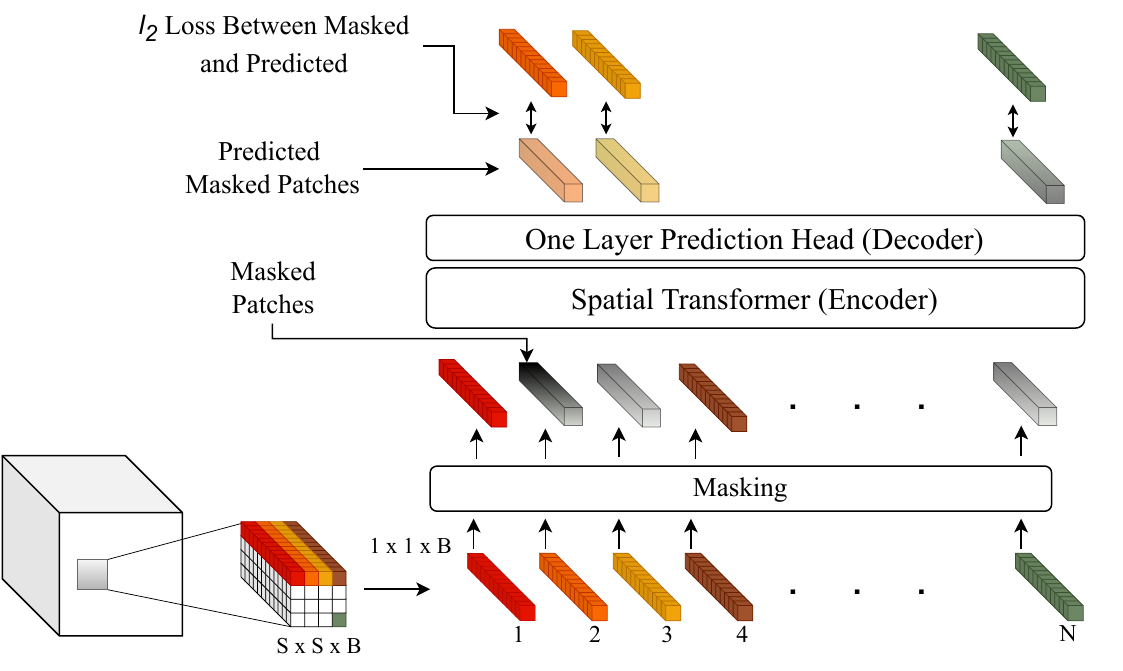}
    \caption{An overview of the proposed pretraining network for spatial transformer in \textit{FactoFormer}.}
    \label{fig:spatialSimMIM}
\end{figure}

\vspace{0.1cm}
 \noindent\textbf{Pretraining of Spatial Transformer:} We apply patching to the input in the spatial dimension and pass them through the spatial transformer along with positional embeddings as outlined in Eq. \ref{eq:spa-proj}. A random proportion of the input tokens are masked, and only the visible tokens are passed through the spatial transformer. Similar to the spectral transformer, we utilize a simple linear layer as the decoder to reconstruct the masked regions and use MSE as the loss function. Notably, the spectrally consistent masks enable the transformer to learn meaningful representations from spatially patched tokens, as depicted in Figure \ref{fig:spatialSimMIM}. 
 
\vspace{0.1cm}
 \noindent\textbf{End-to-End Fine-tuning:} We use the pretrained weights of the spectral and spatial transformers to initialize \textit{FactoFormer}. Then we fine-tune the network end-to-end in order to perform classification. Our proposed factorization technique enables \textit{FactoFormer}  to exploit the pretrained weights and learn rich spectral and spatial feature representations jointly without imposing significant complexity to the network.

%% file: chapters/experiments.tex
\section{Experiments}
\label{sec:exp}
In this section, we present our experimental setup, evaluated datasets, and implementation details of \textit{FactoFormer} and other state-of-the-art methods as set in previous literature \cite{hong2021spectralformer, MAEST}. 

\subsection{Datasets}

We use six popular hyperspectral datasets to evaluate the performance of the proposed model and to compare it with other state-of-the-art methods. These datasets include Indian Pines \cite{IP}, University of Pavia \cite{UP}, and Houston 2013 \cite{Houston2013}, as well as the Wuhan UAV-borne hyperspectral image (WHU-Hi) dataset \cite{wuhan_dataset}, which consists of three individual datasets: WHU-Hi-LongKou, WHU-Hi-HanChuan, and WHU-Hi-HongHu. The details on number of samples used for pretraining, fine-tuning and testing are shown in Table \ref{tab:dataset_details}.

\vspace{0.1cm}
\noindent \textbf{Indian Pines}\cite{IP}
dataset was acquired using an Airborne Visible Infrared Imaging Spectrometer (AVIRIS) in Indiana, USA, in 1992. There are 220 bands in the HSI that cover a wavelength range of 400nm to 2500nm. Upon processing and removing 20 noisy and water absorption bands, the final image consists of 200 spectral bands. The spatial resolution of the image is $145 \times 145$. A total of 16 distinct classes of agriculture and perennial vegetation scenes are captured in this dataset.

\noindent \textbf{Pavia University}\cite{UP} 
dataset was collected in 2001 using the Reflective Optics System Imaging Spectrometer (ROSIS) at Pavia University in Northern Italy. There are 115 spectral bands in the original image ranging from 430 nm to 860 nm with a spatial resolution of $610 \times 340$. The data were prepossessed to remove 12 noisy bands, resulting in 103 bands with nine land-cover classes.

\noindent \textbf{Houston 2013}\cite{Houston2013} dataset was captured using an ITRES CASI-1500 sensor over the University of Houston in Texas, USA. The image contains 144 bands ranging from 364 nm to 1046 nm with a spatial resolution of $349 \times 1905$. The dataset contains fifteen different land cover classes. The dataset was released as a part of the 2013 IEEE Geoscience and Remote Sensing Society (GRSS) Data Fusion contest. 

We have used the same train-test split as in SpectralFormer\cite{hong2021spectralformer} and MAEST\cite{MAEST} for training and evaluating our models on Indian Pines, University of Pavia and Houston 2013 datasets.

\vspace{0.1cm}
\noindent \textbf{WHU-Hi dataset}\cite{wuhan_dataset} comprises three datasets acquired in farming areas with various crop types in Hubei province, China, using a Headwall Nano-Hyperspec sensor for precise crop classification.  For our experiments with WHU-Hi dataset, we followed the same setup described in \cite{wuhan_dataset}.

\noindent \textbf{WHU-Hi-LongKou} dataset was acquired in 2018 in LongKou town using a UAV flown at an altitude of 500m. It consists of 9 classes with six crop species. The dataset has dimensions of $550 \times 400$ pixels and includes 270 spectral bands ranging from 400 nm to 1000 nm. 

\noindent \textbf{WHU-Hi-HanChuan} dataset was collected in HanChuan in 2016 using a UAV flying at an altitude of 250m. The dataset comprises 274 bands within the range of 400 nm to 1000 nm and has a spatial size of $1217 \times 303$. It includes 16 classes with seven different crop species.

\noindent \textbf{WHU-Hi-HongHu} dataset was collected in 2017 using a UAV flying at an altitude of 100m in HongHu city. The dataset contains 22 classes with 17 crop species. The spatial size of the image is $940 \times 475$ and there are 270 bands within the range of 400nm and 1000nm.

\begin{table}[t]
\centering
\caption{Details of Data used for pretraining, Fine-tuning and Testing}
\label{tab:dataset_details}
\begin{tabular}{cccc}
\hline 
Dataset & Pre-train  & Fine-tune  & Test   \\
\hline 
Indian Pines & 10659 & 695 & 9671  \\
University of Pavia  & 163477 & 3921 & 40002 \\
Houston 2013 & 649816 & 2832 & 12197 \\
WHU-Hi-LongKou & 15458 & 900 & 203642 \\
WHU-Hi-HanChuan & 111221 & 1600 & 255930 \\
WHU-Hi-HongHu & 59807 & 2200 & 384493 \\
\hline
\end{tabular}
\end{table}

\subsection{Implementation Details}

\noindent \textbf{Network Architectures:} We use a transformer architecture with five layers and four attention heads for both spectral and spatial transformers.%
We set the embedding size for spectral and spatial transformers as 32 and 64, respectively. Table \ref{tab:vit_details} outlines the details of the network architectures. The same architectures were preserved during pretraining as well.  

\begin{table}[b]
\caption{Details of Spectral and Spatial Transformer Networks}
\label{tab:vit_details}
\begin{tabular}{c|c c c c}
\hline 
        Model         & Layers &  \begin{tabular}[c]{@{}c@{}}Embedding \\ Size\end{tabular} & MLP size & Heads \\
\hline
Spectral Transformer & 5      & 32             & 4        & 4  \\
Spatial Transformer  & 5      & 64             & 8        & 4     \\
\hline
\end{tabular}
\end{table}

\vspace{0.1cm}
\noindent \textbf{Pretraining:}  We employ the Adam optimizer with a linear learning rate scheduler and train for 200 epochs. The batch size is set to 32 with an initial learning rate of $5e^{-4}$ and decayed by a factor of 0.9 for every 20 epochs. The pretraining setting is detailed in Table \ref{tab:pretraining_setup}.

\vspace{0.1cm}
\noindent \textbf{Fine-tuning:}  For fine-tuning, the pretrained weights are used to initialize both spectral and spatial transformers and fine-tuned end-to-end. A learning rate search is performed, and the best for each dataset is reported. We use the Adam optimizer with a linear learning rate scheduler and set the batch size to 32. For both the Indian Pines and University of Pavia datasets, the fine-tuning model converges after 80 epochs, while for Houston and WHU-Hi datasets, it converges in just 40 epochs. The Finetuning setup is detailed in Table \ref{tab:finetuning_setup}.

\begin{table}[t]
    \begin{minipage}[t]{.5\linewidth}
        \centering
        \caption{Pretraining Setting}
        \label{tab:pretraining_setup}
        \begin{tabular}{l|l}
        \hline
        config & Value \\
        \hline 
        Optimizer & Adam \\
        learning rate & $5e-4$ \\
        learning rate schedule & StepLR \\
        weight decay & 0 \\
        batch size & 32 \\
        epochs & 200 \\
        masking ratio & 0.7 \\
        \hline
        \end{tabular}
    \end{minipage}%
    \begin{minipage}[t]{.5\linewidth}
      \centering
        \caption{Fine-tuning Setting}
        \label{tab:finetuning_setup}
        \begin{tabular}{l|l}
        \hline
        config & Value \\
        \hline 
        Optimizer & Adam \\
        learning rate schedule & StepLR \\
        weight decay & 0 \\
        batch size & 32 \\
        \hline
        learning rate &  \\
        Indian Pines & $3e-4$ \\
        University of Pavia & $1e-2$ \\
        Houston 2013 & $2e-3$ \\
        WHU-Hi datasets & $1e-3$ \\ 
        \hline
        epochs &  \\
        Indian Pines & 80 \\
        University of Pavia & 80 \\
        Houston 2013 & 40 \\
        WHU-Hi datasets & 40 \\ 
        \hline
        \end{tabular}
    \end{minipage} 
\end{table}

\vspace{0.1cm}
\noindent\textbf{Evaluation Matrices}: For evaluating the classification performance quantitatively, we use the three generally used matrices Overall Accuracy (OA), Average Accuracy (AA), and Kappa Coefficient (k).

\subsection{State-of-the-art Methods}
\noindent\textbf{Conventional Classifiers}: We used KNN\cite{knn-song2016hyperspectral}, SVM\cite{SVM}, and RF\cite{RF} as conventional classification methods. The number of nearest neighbors (K) is set to 10 for KNN\cite{knn-song2016hyperspectral}. SVM\cite{SVM} was implemented using a radial basis function (RBF) kernel. The main two hyperparameters of RBF kernel $\sigma$ and $\lambda$ are optimized using fivefold cross-validation on the training set with the range of $\sigma = \left [ 2^{-3}, 2^{-2}, ..., 2^{4}\right ] $ and $\lambda = \left [ 10^{-2}, 10^{-1}, ..., 10^{4}\right ] $ respectively. In the Random Forest method (RF)\cite{RF}, 200 decision trees are used for experiments. 

\vspace{0.1cm}
\noindent\textbf{Classic Backbone Networks}: We compare our method with 1-D-CNN\cite{1D-CNN}, 2-D CNN\cite{2D-CNN}, RNN\cite{mou2017deep}, and miniGCN\cite{miniGCN} methods. For the 1D-CNN model\cite{1D-CNN}, 1-D convolutional filters with output sizes of 128 are combined with a batch normalization layer and ReLU activation function. Finally, a Softmax function is applied to the last layer of the model for classification. The 2-D CNN\cite{2D-CNN} comprises three blocks, each consisting of a 2-D convolutional layer, a batch normalization layer, and a max-pooling layer with ReLU activation. In each block, convolutional kernels with filter sizes of $3 \times 3 \times 32, 3 \times 3 \times 64 $, and $1 \times 1 \times 128$ are utilized. RNN\cite{mou2017deep} is implemented using two recurrent layers with gated recurrent units of 128 neurons in each layer. In miniGCN\cite{miniGCN}, the network block consists of a batch normalization layer, a graph convolutional layer with 128 neurons, and a ReLU layer.   

\vspace{0.1cm}
\noindent\textbf{Transformer Based Networks}: For transformer-based networks, we used the ViT architecture \cite{vit_original}, SpectralFormer \cite{hong2021spectralformer}, and MAEST \cite{MAEST}. The ViT uses five layers, four attention heads, an embedding size of 64, and an MLP with a hidden dimension of 8. SpectralFormer\cite{hong2021spectralformer} uses the same settings as the ViT with cross-layer adaptive fusion (CAF). CAF employs a skip connection mechanism to integrate the features of skipping encoder blocks. MAEST\cite{MAEST} is a self-supervised approach that utilizes Masked Autoencoders (MAE) to train an encoder-decoder network that reconstructs masked patches. Then a classification encoder with similar network architecture as SpectralFormer\cite{hong2021spectralformer} is initialized with the pretrained encoder weights and fine-tuned to perform classification.

%% file: chapters/results.tex
\section{Results}
\label{sec:results}
\begin{figure*}[t]
    \centering
    \includegraphics[scale=0.9]{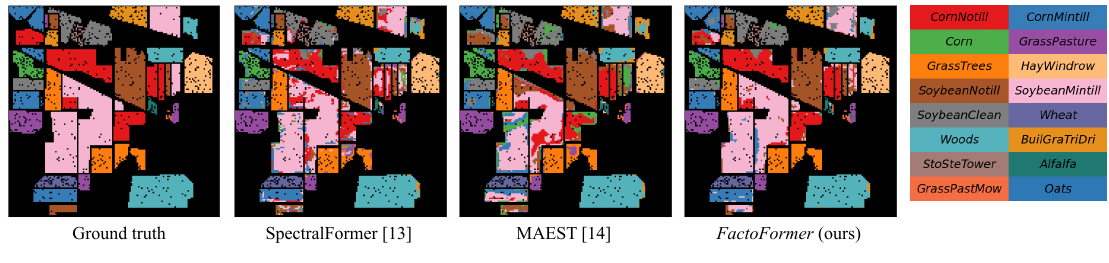}
    \caption{Ground truth and the classification maps obtained by different models on the Indian Pines dataset.}
    \label{fig:cls_map_ip}
\end{figure*}

\begin{figure*}[t]
    \centering
    \includegraphics[scale=0.8]{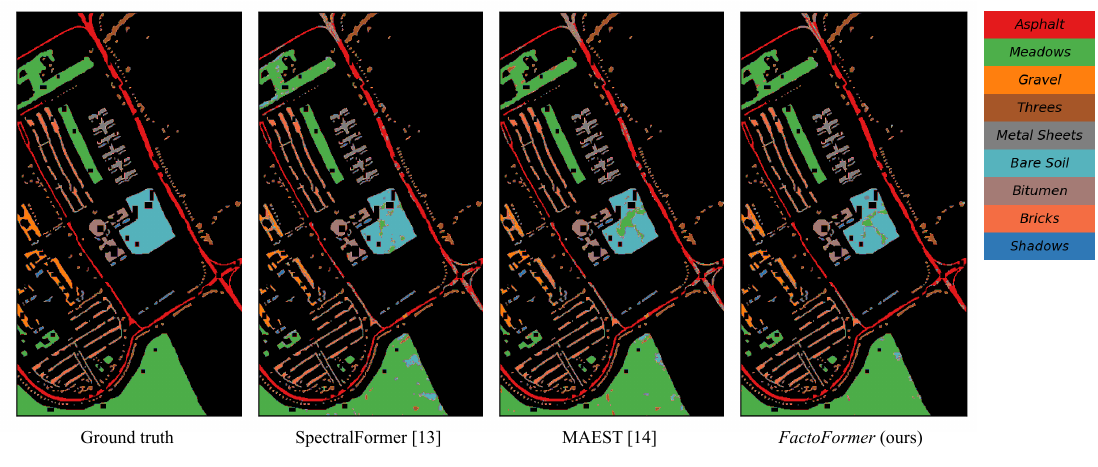}
    \caption{Ground truth and the classification maps obtained by different models on the University of Pavia dataset.}
    \label{fig:cls_map_up}
\end{figure*}

\begin{figure*}[h]
    \centering
    \includegraphics[scale=0.7]{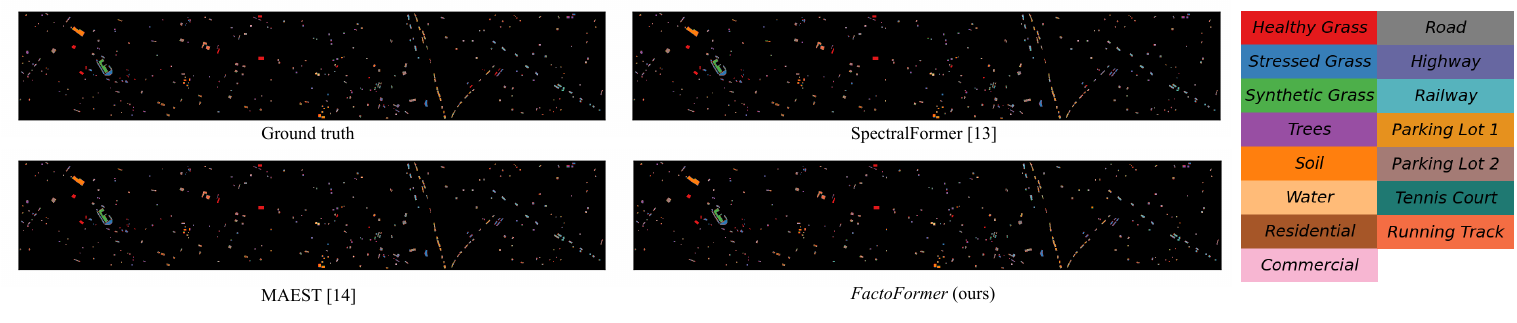}
    \caption{Ground truth and the classification maps obtained by different models on the Houston 2013 dataset.}
    \label{fig:cls_map_houston}
\end{figure*}

We compare our proposed \textit{FactoFormer} architecture with state-of-the-art transformer networks, classic backbone networks, and conventional machine learning approaches. These evaluations are conducted on Indian Pines, University of Pavia, and Houston 2013 HSI datasets.  We used OA, AA, and Kappa coefficient as the evaluation matrices following the convention, and the results are shown in Table \ref{tab:IP}, \ref{tab:UP} and \ref{tab:Houston} along with class-wise accuracy. Also, we compare the classification maps with ground truth and other transformer based models as shown in Figure \ref{fig:cls_map_ip}, \ref{fig:cls_map_up} and \ref{fig:cls_map_houston}. Further we evaluate our method on WHU-Hi dataset which contains three challenging UAV-borne datasets and compare it with other state-of-the-art transformer based approaches and the results are shown in Table \ref{tab:wuhan_results}. \textit{FactoFormer} achieve state-of-the-art performance in all datasets with significant improvements.

\begin{table*}[t]
\centering
\caption{Quantitative Performance of Different Classification Methods in Terms of OA, AA, and \textit{k} as Well as the Accuracy for Each Class on the Indian Pines Dataset. The Best One is Shown in Bold}
\label{tab:IP}
\resizebox{\textwidth}{!}{\begin{tabular}{c|ccc|cccc|c|cc|cc|c}
\hline 
Class & \multicolumn{3}{c|}{Conventional Classifiers*} & \multicolumn{4}{c|}{Classic Backbone Networks*} & \multicolumn{6}{c}{Transformers} \\
\hline

\multicolumn{1}{c|}{} & \multicolumn{1}{c}{KNN\cite{knn-song2016hyperspectral}} & \multicolumn{1}{c}{RF\cite{RF}} & \multicolumn{1}{c|}{SVM\cite{SVM}} &\multicolumn{1}{c}{1-D CNN\cite{1D-CNN}} &\multicolumn{1}{c}{2-D CNN\cite{2D-CNN}} &\multicolumn{1}{c}{RNN\cite{mou2017deep}} & \multicolumn{1}{c|}{miniGCN \cite{miniGCN}} &\multicolumn{1}{c|}{ ViT*\cite{vit_original}} & \multicolumn{2}{c|}{SpectralFormer \cite{hong2021spectralformer}} & \multicolumn{2}{c|}{MAEST \cite{MAEST}} & \multicolumn{1}{c}{ \textit{FactoFormer} (ours) }\\

 &  &  &  & & & &  & & pixel & \multicolumn{1}{c|}{patch} & \multicolumn{1}{c}{pixel} & patch & \\
 \hline
Corn Notil & 45.45 & 57.80 & 67.34 & 47.83 & 65.90 & 69.00 & 72.54 & 53.25 & 62.64 & 70.52 & 68.64 & 78.97 & \textbf{85.98} \\
Corn Mintill & 46.94 & 56.51 & 67.86 & 42.35 & 76.66 & 58.93 & 55.99 & 66.20 & 66.20 & 81.89 & 76.91 & \bf92.73 &  92.35\\
Corn & 77.72 & 80.98 & 93.48 & 60.87 & 92.39 & 77.17 & 92.93 & 86.41 & 88.59 & 91.30 & 97.28 & \bf98.91 & 95.65\\
Grass Pasture & 84.56 & 85.68 & 94.63 & 89.49 & 93.96 & 82.33 & 92.62 & 89.71 & 90.16 & 95.53 & 93.51 & 96.20 & \textbf{96.64}\\
Grass Trees & 80.06 & 79.34 & 88.52 & 92.40 & 87.23 & 67.72 & 94.98 & 87.66 & 89.24 & 85.51 & 87.52 & 89.10 &  \textbf{100.0}\\
Hay Windrowed & 97.49 & 95.44 & 94.76 & 97.04 & 97.27 & 89.07 & 98.63 & 89.98 & 95.90 & 99.32 & 89.98 & 98.41 &  \textbf{99.77}\\
Soybean Notill & 64.81 & 77.56 & 73.86 & 59.69 & 77.23 & 69.06 & 64.71 & 72.22 & 85.19 & 81.81 & 84.64 & 84.86 &  \textbf{88.67}\\
Soybean Mintill & 48.68 & 58.85 & 52.07 & 65.38 & 57.03 & 63.56 & 68.78 & 66.00 & 74.48 & 75.48 & 69.44 & 73.44 &  \textbf{85.57}\\
Soybean Clean & 44.33 & 62.23 & 72.70 & \bf93.44 & 72.87 & 65.07 & 69.33 & 57.09 & 72.34 & 73.76 & 74.11 & 69.50 &  91.13\\
Wheat & 96.30 & 95.06 & 98.77 & 99.38 & \bf100.00 & 95.06 & 98.77 & 97.53 & 98.15 & 98.77 & 98.15 & \bf100.00 & 99.38 \\
Woods & 74.28 & 88.75 & 86.17 & 84.00 & 92.85 & 88.67 & 87.78 & 87.62 & 93.01 & 93.17 & 84.97 & 92.12 &  \textbf{96.54}\\
Buildiing Grass Trees Drives & 15.45 & 54.24 & 71.82 & 86.06 & 88.18 & 50.00 & 50.00 & 63.94 & 60.91 & 78.48 & 76.36 & 91.51 &  \textbf{97.58}\\
Stone Steel Towers & 91.11 & 97.78 & 95.56 & 91.11 & \bf100.00 & 97.78 & \bf100.00 & 95.56 & \bf100.00 & \bf100.00 & \bf100.0 & \bf100.00 &  \textbf{100.00}\\
Alfalfa & 33.33 & 56.41 & 82.05 & 84.62 & 84.62 & 66.67 & 48.72 & 79.49 & 87.18 & 79.49 & \textbf{94.87} & 89.74 & 79.87\\
Grass Pasture Mowed & 81.82 & 81.82 & 90.91 & \bf100.00 & \bf100.00 & 81.82 & 72.73 & 90.91 & 90.91 & \bf100.00 & 90.90 & \bf100.00&  \textbf{100.00} \\
Oats & 40.00 & \bf100.00 & \bf100.00 & 80.00 & \bf100.00 & \bf100.00 & 80.00 & 80.00 & \bf100.00 & \bf100.00 & \bf100.00 & \bf100.00 & \textbf{100.00} \\
\hline 
OA (\%) & 59.17 & 69.8 & 72.36 & 70.43 & 75.89 & 70.66 & 75.11 & 71.86 & 78.55 & 81.76 & 78.52 & 84.15 & \textbf{91.30} \\
AA (\%) & 63.9 & 76.78 & 83.16 & 79.6 & 86.64 & 76.37 & 78.03 & 78.97 & 84.68 & 87.81 & 86.71 & 90.97 &  \textbf{94.30}\\
\textit{k} & 0.5395 & 0.6591 & 0.6888 & 0.6642 & 0.7281 & 0.6673 & 0.7164 & 0.6804 & 0.7554 & 0.7919 & 0.7567 & 0.820 & \textbf{0.9006} \\
\hline 
\multicolumn{12}{l}{*Reports the results from the SpectralFormer paper \cite{hong2021spectralformer}.}
\end{tabular}}
\end{table*}

\begin{table*}[t]
\centering
\caption{Quantitative Performance of Different Classification Methods in Terms of OA, AA, and \textit{k} as Well as the Accuracy for Each Class on the University of Pavia Dataset. The Best One is Shown in Bold.}
\label{tab:UP}
\resizebox{\textwidth}{!}{\begin{tabular}{c|ccc|cccc|c|cc|cc|c}
\hline 
Class  & \multicolumn{3}{c|}{Conventional Classifiers*} & \multicolumn{4}{c|}{Classic Backbone Networks*} & \multicolumn{6}{c}{Transformers} \\
\hline
\multicolumn{1}{c|}{} & \multicolumn{1}{c}{KNN\cite{knn-song2016hyperspectral}} & \multicolumn{1}{c}{RF\cite{RF}} & \multicolumn{1}{c|}{SVM\cite{SVM}} &\multicolumn{1}{c}{1-D CNN\cite{1D-CNN}} &\multicolumn{1}{c}{2-D CNN\cite{2D-CNN}} &\multicolumn{1}{c}{RNN\cite{mou2017deep}} & \multicolumn{1}{c|}{miniGCN \cite{miniGCN}} &\multicolumn{1}{c|}{ ViT*\cite{vit_original}} & \multicolumn{2}{c|}{SpectralFormer \cite{hong2021spectralformer}} & \multicolumn{2}{c|}{MAEST \cite{MAEST}} & \multicolumn{1}{c}{ \textit{FactoFormer} (ours)}\\

 &  &  &  & & & &  & & pixel & \multicolumn{1}{c|}{patch} & \multicolumn{1}{c}{pixel} & patch & \\
 \hline 
Asphalt & 73.86 & 79.81 & 74.22 & 88.90 & 80.98 & 84.01 & \textbf{96.35} & 71.51 & 82.95 & 82.73 & 85.45 & 79.85 & 92.04 \\
Meadows & 64.31 & 54.90 & 52.79 & 58.81 & 81.70 & 66.95 & 89.43 & 76.82 & 95.23 & 94.03 & 77.24 & 96.60 & \textbf{98.64} \\
Gravel & 55.10 & 46.34 & 65.45 & 73.11 & 67.99 & 58.46 & \bf87.01 & 46.39 & 78.18 & 73.66 & 68.60 & 67.66 & 75.76 \\
Trees & 94.95 & \bf98.73 & 97.42 & 82.07 & 97.36 & 97.70 & 94.26 & 96.39 & 87.95 & 93.75 & 97.49 & 96.84 & 95.30 \\
Metal Sheets & 99.19 & 99.01 & 99.46 & 99.46 & 99.64 & 99.10 & \bf99.82 & 99.19 & 99.46 & 99.28 & 99.28 & 99.73 & 98.11 \\
Bare Soil & 65.16 & 75.94 & 93.48 & \bf97.92 & 97.59 & 83.18 & 43.12 & 83.18 & 65.84 & 90.75 & 90.94 & 82.70 & 89.39 \\
Bitumen & 84.30 & 78.70 & 87.87 & 88.07 & 82.47 & 83.08 & 90.96 & 83.08 & 92.35 & 87.56 & 89.09 & 96.23 & \textbf{98.78} \\
Bricks & 84.10 & 90.22 & 89.39 & 88.14 & 97.62 & 89.63 & 77.42 & 89.63 & 85.26 & 95.81 & 91.20 & 95.96 & \textbf{98.51} \\
Shadows & 98.36 & 97.99 & 99.87 & 99.87 & 95.60 & 96.48 & 87.27 & 96.48 & \bf100.00 & 94.21 & 99.50 & 94.47 & 96.23 \\
\hline 
OA (\%) & 70.53 & 69.67 & 70.82 & 75.5 & 86.05 & 77.13 & 79.79 & 76.99 & 87.94 & 91.07 & 83.70 & 91.06 & \textbf{95.19} \\
AA (\%) & 79.68 & 80.18 & 84.44 & 86.26 & 88.99 & 84.29 & 85.07 & 80.22 & 87.47 & 90.2 & 88.76 & 90.00 & \textbf{93.64} \\
\textit{k} & 0.6268 & 0.6237 & 0.6423 & 0.6948 & 0.8187 & 0.7101 & 0.7367 & 0.701 & 0.8358 & 0.8805 & 0.7899 & 0.8794 & \textbf{0.9349} \\
\hline 
\multicolumn{12}{l}{*Reports the results from the SpectralFormer paper \cite{hong2021spectralformer}.}
\end{tabular}}
\end{table*}

\vspace{0.1cm}
\begin{table*}[t]
\centering
\caption{Quantitative Performance of Different Classification Methods in Terms of OA, AA, and \textit{k} as Well as the Accuracy for Each Class on the Houston 2013 Dataset. The Best One is Shown in Bold}
\label{tab:Houston}
\resizebox{\textwidth}{!}{\begin{tabular}{c|ccc|cccc|c|cc|cc|c}
\hline 
Class  & \multicolumn{3}{c|}{Conventional Classifiers*} & \multicolumn{4}{c|}{Classic Backbone Networks*} & \multicolumn{6}{c}{Transformers} \\
\hline
\multicolumn{1}{c|}{} & \multicolumn{1}{c}{KNN\cite{knn-song2016hyperspectral}} & \multicolumn{1}{c}{RF\cite{RF}} & \multicolumn{1}{c|}{SVM\cite{SVM}} &\multicolumn{1}{c}{1-D CNN\cite{1D-CNN}} &\multicolumn{1}{c}{2-D CNN\cite{2D-CNN}} &\multicolumn{1}{c}{RNN\cite{mou2017deep}} & \multicolumn{1}{c|}{miniGCN \cite{miniGCN}} &\multicolumn{1}{c|}{ ViT*\cite{vit_original}} & \multicolumn{2}{c|}{SpectralFormer \cite{hong2021spectralformer}} & \multicolumn{2}{c|}{MAEST \cite{MAEST}} & \multicolumn{1}{c}{ \textit{FactoFormer} (ours) }\\

 &  &  &  & & & &  & & pixel & \multicolumn{1}{c|}{patch} & \multicolumn{1}{c}{pixel} & patch & \\
 \hline 
Healthy Grass & 83.19 &	83.38 &	83.00 & 87.27 &	85.09 &	82.34 &	\bf98.39 &	82.81 &	83.48 &	81.86 & 85.00 & 82.90 & 89.13\\
Stressed Grass & 95.68 & 98.40 & 98.40 & 98.21 & 99.91 & 94.27 & 92.11 & 96.62 & 95.58 & \bf100.00 & 98.40 & 99.44 & 99.62  \\
Synthetic Grass & 99.41 & 98.02 & 99.60 & \bf100.00 & 77.23 & 99.60 & 99.60 & 99.80 & 99.60 & 95.25 & 99.80 & 94.65 & 97.43 \\
Tree & 97.92 & 97.54 & 98.48 & 92.99 & 97.73 & 97.54 & 96.78 & \bf99.24 & 99.15 & 96.12 & 96.50 & 96.02 & 95.08 \\
Soil & 96.12 & 96.40 & 97.82 & 97.35 & \textbf{99.53} & 93.28 & 97.73 & 97.73 & 97.44 & \textbf{99.53} & 98.39 & 99.24 & 98.866 \\
Water & 92.31 & \bf97.20 & 90.91 & 95.10 & 92.31 & 95.10 & 95.10 & 95.10 & 95.10 & 94.41 & 95.10 & 93.01 & 93.01 \\
Residential & 80.88 & 82.09 & 90.39 & 77.33 & \bf92.16 & 83.77 & 57.28 & 76.77 & 88.99 & 83.12 & 88.53 & 89.93 & 84.79 \\
Commercial & 48.62 & 40.65 & 40.46 & 51.38 & 79.39 & 56.03 & 68.09 & 55.65 & 73.31 & 76.73 & 57.64 & \bf82.15 & 78.54 \\
Road & 72.05 & 69.78 & 41.93 & 27.95 & \bf86.31 & 72.14 & 53.92 & 67.42 & 71.86 & 79.32 & 69.59 & 77.15 & 79.32 \\
Highway & 53.19 & 57.63 & 62.64 & \bf90.83 & 43.73 & 84.17 & 77.41 & 68.05 & 87.93 & 78.86 & 89.29 & 87.55 & 86.68 \\
Railway & 86.24 & 76.09 & 75.43 & 79.32 & 87.00 & 82.83 & 84.91 & 82.35 & 80.36 & 88.71 & 88.52 & 85.58 & \textbf{95.54} \\
Parking Lot 1 & 44.48 & 49.38 & 60.04 & 76.56 & 66.28 & 70.61 & 77.23 & 58.50 & 70.70 & \bf87.32 & 73.78 & 80.60 & 82.80 \\
Parking Lot 2 & 28.42 & 61.40 & 49.47 & 69.47 & \bf90.18 & 69.12 & 50.88 & 60.00 & 71.23 & 72.63 & 71.23 & 69.62 & 88.77 \\
Tennis Court & 97.57 & 99.60 & 98.79 & 99.19 & 90.69 & 98.79 & 98.38 & 98.79 & 98.79 & \bf100.00 & \bf100.00 & 97.17 & 98.79 \\
Running Track & 98.10 & 97.67 & 97.46 & 98.10 & 77.80 & 95.98 & 98.52 & 98.73 & 98.73 & \bf99.79 & 97.89 & 98.10 & 89.43 \\
\hline 
OA (\%) & 77.30 & 77.48 & 76.91 & 80.04 & 83.72 & 83.23 & 81.71 & 80.41 & 86.14 & 88.01 & 85.86 & 88.55 & \textbf{89.13} \\
AA (\%) & 78.28 & 80.35 & 78.99 & 82.74 & 84.35 & 85.04 & 83.09 & 82.50 & 87.48 & 88.91 & 87.31 & 88.89 & \textbf{90.12} \\
\textit{k} & 0.7538 & 0.7564 & 74.94 & 0.7835 & 0.8231 & 0.8183 & 0.8018 & 0.7876 & 0.8497 & 0.8699 & 0.8467 & 0.8757 & \textbf{0.8820} \\
\hline 
 \multicolumn{12}{l}{*Reports the results from the SpectralFormer paper \cite{hong2021spectralformer}.}
\end{tabular}}
\end{table*}

\begin{table}[]
\centering
\caption{Comparison of FactoFormer With Transformer based Methods on WHU-Hi Dataset}
\label{tab:wuhan_results}
\resizebox{\columnwidth}{!}{\begin{tabular}{c|c||ccc}
\hline 
Dataset & Metric & \begin{tabular}[c]{@{}c@{}} SpectralFormer* \cite{hong2021spectralformer} \\\end{tabular} & \begin{tabular}[c]{@{}c@{}} MAEST* \cite{MAEST} \end{tabular} & \begin{tabular}[c]{@{}c@{}} \textit{FactoFormer}\\ (ours)\\ \end{tabular} \\
\hline
& OA (\%) & 93.30 & 94.34 & \bf98.30 \\
WHU-Hi- & AA (\%) & 95.46 & 94.40 &  \bf98.72 \\
LongKou& \textit{k} & 0.9137 & 0.9266 & \bf0.9778 \\
\hline
& OA (\%) & 89.85 & 84.85 & \bf93.19 \\
WHU-Hi- & AA (\%) & 88.31 & 81.77 & \bf91.64 \\
HanChuan & \textit{k} & 0.8816 & 0.8239 &  \bf0.9205\\
\hline
& OA (\%) & 90.16 & 86.93 & \bf92.26\\
WHU-Hi- & AA (\%) & 89.11 & 84.09 & \bf92.38 \\
HongHu & \textit{k} & 0.8768 & 0.8366 & \bf0.9029 \\
\hline 
\multicolumn{5}{l}{* For a fair comparison, we used the official repository of SpectralFormer\cite{hong2021spectralformer} and}\\
\multicolumn{5}{l}{MAEST\cite{MAEST} to produce the results on WHU-Hi dataset.}\\
\end{tabular}}
\end{table}

\subsection{Comparison with the State-of-the-Art}
When comparing \textit{FactoFormer} with MAEST\cite{MAEST}, which is the current state-of-the-art method using a transformer with a self-supervised pretraining approach, our method outperforms it with 7.15\%, 4.13\%, and 0.58\% higher OA on Indian Pines, University of Pavia and Houston 2013 datasets respectively. It shows that our novel factorized approach with self-supervised pretraining improves performance significantly. Also, interestingly our models converge with less than 100 epochs during fine-tuning on all three datasets, whereas, in MAEST\cite{MAEST}, the models are fine-tuned for more than 200 epochs. It again highlights that the factorized pretraining in \textit{FactoFormer} learns rich feature representations, which enables fast convergence during fine-tuning. Also, \textit{FactoFormer} is computationally more efficient in terms of the number of model parameters and run-time, which is briefly discussed in Section \ref{sec:ablations} under computational efficiency.

Compared to SpectralFormer \cite{hong2021spectralformer}, which is a fully supervised transformer-based approach, \textit{FactoFormer} outperforms it by 9.54\% on Indian Pines, 4.12\% on University of Pavia and 1.12\% on Houston 2013 datasets highlighting the importance of factorization and pretraining. ViT\cite{vit_original} shows the least performance of all the transformer methods, as it is applied directly as in images for spectral tokens in HSI without adapting it to suit the structure of hyperspectral data. In general, SpectralFormer\cite{hong2021spectralformer}, MAEST\cite{MAEST}, and \textit{FactoFormer} perform better than all other classic backbone networks and conventional classifiers in all three datasets, demonstrating the effectiveness of transformer-based approaches for HSI. 2-D CNN\cite{2D-CNN} performs better compared to 1-D CNN\cite{1D-CNN}, RNN\cite{mou2017deep} and miniGCN\cite{miniGCN} on all three datasets. When comparing 2-D CNN \cite{2D-CNN}, \textit{FactoFormer} achieves 15.41\%, 9.14\%, and 5.41\% higher accuracy on Indian Pines, University of Pavia and Houston 2013 datasets, respectively. This significant performance gap exhibits the effectiveness of \textit{FactoFormer} over CNN-based approaches.

\vspace{0.2cm}
In order to demonstrate the robust performance of \textit{FactoFormer} across more challenging datasets, we conducted experiments with the WHU-HI dataset, which comprises three UAV-borne datasets. We compared the performance against previous state-of-the-art transformer-based methods. \textit{FactoFormer} achieved 3.96\%, 8.34\%, and 5.33\% higher overall accuracy than MAEST \cite{MAEST} and 5\%, 3.34\%, and 2.1\% higher overall accuracy than SpectralFormer \cite{hong2021spectralformer} on the WHU-Hi-LongKou, WHU-Hi-HanChuan, and WHU-Hi-HongHu datasets, respectively. These three datasets have high spectral resolution with around 270 bands, which is crucial for learning and distinguishing subtle differences between different crop species. The performance improvement with our method demonstrates that utilizing both spatial and spectral features in \textit{FactoFormer} is important, especially with downstream tasks like precise crop classification, where variations in spectra among different species are not very significant. This stands in contrast to the approach of focusing only on spectra, as seen in SpectralFormer \cite{hong2021spectralformer} and MAEST \cite{MAEST}.

\subsection{Effect of Factorization}

In order to showcase the significance of factorized transformers and underscore their advantages in pretraining, we compare the performance with pretraining and without pretraining, \ie training from scratch in a fully supervised manner, for both \textit{FactoFormer} and MAEST\cite{MAEST}. For this, we trained \textit{FactoFormer} with a fully supervised setup by training the factorized transformers end-to-end from scratch for 300 epochs while searching for the best hyper-parameters. Accuracy of SpectralFormer\cite{hong2021spectralformer} is considered as non-pretrained MAEST\cite{MAEST} since the architecture of the classification encoder in MAEST\cite{MAEST} is the same as SpectralFormer\cite{hong2021spectralformer}. Table \ref{tab:factoformervsMAEST} shows the results evaluated on the Indian Pines dataset, and OA, AA and k are reported. \textit{FactoFormer} outperforming MAEST\cite{MAEST} without pretraining shows that factorization facilitates learning more meaningful feature representations from both spectral and spatial dimensions. When we compare the improvement of accuracy when pretrained, MAEST\cite{MAEST} shows a 2.39\% improvement, whereas \textit{FactoFormer} improves by 8.67\%. This significant improvement shows that factorization enables more meaningful and effective pretraining. Moreover, pretrained, \textit{FactoFormer} outperforming pretrained MAEST \cite{MAEST} by 7.15\% proves the effectiveness of \textit{FactoFormer} for hyperspectral data.    

\begin{table}[]
\centering
\caption{Effectiveness of Factorization in Supervised and Self-supervised Setups}
\label{tab:factoformervsMAEST}
\begin{tabular}{c|c||cc}
\hline 
Setup & Metric & MAEST\cite{MAEST} & \textit{FactoFormer} (ours)  \\
\hline
 & OA (\%) & 81.76 & \textbf{82.63} \\
w/o Pre-training & AA (\%) & 87.81 &  \textbf{89.11} \\
 & k & 0.7919 & \textbf{0.8029} \\
 \hline
& OA (\%) &  84.15 &  \textbf{91.30}\\
w/ Pre-training  & AA (\%) & 90.97 & \textbf{94.30} \\
 & \textit{k} & 0.8200 & \textbf{0.9006}   \\
\hline 
\end{tabular}
\end{table}

\subsection{Ablation Study}
\label{sec:ablations}
In this section, we conduct ablation studies to analyze the contribution of each component in \textit{FactoFormer}. First, we investigate the advantage of using the factorized architecture over using them separately. Next, we examine the efficiency of using self-supervised pretraining compared to a fully supervised setting. Then we attempt pretraining with different masking ratios and different patch sizes. Finally, we analyze the computational efficiency of \textit{FactoFormer}. Unless stated otherwise, we use the Indian Pines dataset for all our experiments.  

\vspace{0.1cm}
\noindent \textbf{Factorized Architecture:} To showcase the effectiveness of the factorized transformers, we conducted separate pretraining and fine-tuning for each transformer. As presented in Table \ref{tab:different_architectures}, a significant drop in performance was observed when only one transformer was used. The factorized architecture allows the spectral and spatial transformers to learn feature representations through spatially and spectrally consistent masking strategies. When fine-tuning \textit{FactoFormer} end-to-end, both transformers are initialized with separately pretrained weights, which allows the model to benefit from the complementary features extracted from both transformers, leading to improved performance.

\begin{table}[t]
\centering
\caption{Comparison of FactoFormer With Spectral and Spatial Transformer}
\label{tab:different_architectures}
\begin{tabularx}{\columnwidth}{c|c||ccc}
\hline 
Dataset & Metric & \begin{tabular}[c]{@{}c@{}}Spectral \\ Transformer\end{tabular} & \begin{tabular}[c]{@{}c@{}} Spatial \\ Transformer\end{tabular} & \begin{tabular}[c]{@{}c@{}} \textit{FactoFormer}\\ (ours)\\ \end{tabular} \\
\hline
& OA (\%) & 80.27 & 86.87 & \bf 91.30 \\
Indian Pines & AA (\%) & 87.73 & 92.55 & \bf 94.30 \\
& \textit{k} & 0.7756 & 0.8505 & \bf 0.9006 \\
\hline
& OA (\%) & 82.51 & 90.38 & \bf 95.19 \\
Univeristy of & AA (\%) & 81.97 & 88.22 & \bf 93.69 \\
Pavia & \textit{k} & 0.7649 & 0.8687 & \bf 0.9349 \\
\hline
& OA (\%) & 81.30 & 87.30 & \bf 89.13 \\
Houston2013 & AA (\%) & 82.62 & 89.48 & \bf 90.12 \\
& \textit{k} & 0.7971 & 0.8623 & \bf 0.8820 \\
\hline 
\end{tabularx}
\end{table}

\begin{figure}[t]
    \centering
    \includegraphics[scale=0.5]{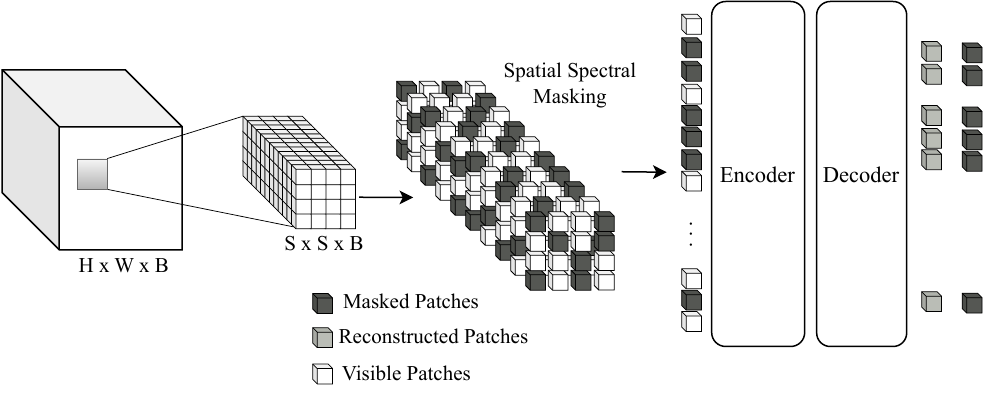}
    \caption{An overview of masked spectral-spatial transformer which is pretrained by masking random portion of spectral-spatial patches and reconstructing the masked regions.}
    \label{fig:jsst}
\end{figure}

\noindent \textbf{Comparison to Masked Spatial-Spectral Transformer:} To highlight the performance and the computational efficiently of \textit{FactoFormer} we compare it with a masked spatial-spectral transformer. Initially we pretrain the model by masking spectral-spatial patches and reconstructing them as shown in Figure \ref{fig:jsst}. We conducted the experiment by patching the input sample $S \times S \times B$ into non overlapping patches of size $1 \times 1 \times k$. Then the pretrained encoder is used to fine-tune the network to perform HSI classification. We set $k$ to $10$ and conducted experiments using the Indian Pines dataset finding the best hyper-parameter setup. As shown in Table \ref{tab:comp-complexity-jsst}, \textit{FactoFormer} performs significantly better with high accuracy and less than half the FLOPs when compared to the masked spatial-spectral transformer network.

\begin{table}[b]
    \centering
    \caption{Computational complexity analysis of FactoFormer with Joint Spectral-spectral transformer}
    \label{tab:comp-complexity-jsst}
    \begin{tabular}{l||cc}
    \hline
     &  \begin{tabular}[c]{@{}c@{}} Overall Accuracy\\  (\%)\end{tabular} &  \begin{tabular}[c]{@{}c@{}} FLOPs \\  (M)\end{tabular} \\
     \hline
     \textit{FactoFormer} (ours)  & 91.30 & 10.77 \\
     Masked Spatial-Spectral Transformer  &  76.22 &  23.17 \\
    \hline
    \end{tabular}
\end{table}

\vspace{0.1cm}
\noindent \textbf{Masking Ratios:} The masking ratio determines the number of visible tokens during pretraining, which is crucial for the model to learn and understand the patterns in the input data. However, setting this value too high or too low can negatively affect performance. In order to determine the optimal masking ratio for pretraining, we conducted a series of experiments using different masking ratios (0.5 to 0.8) for both spectral and spatial transformers. The results of this grid search are presented in Table \ref{tab:ratios}. Based on this study, we selected a masking ratio of 0.7 for both transformers as the optimal value for fine-tuning the final model.

\begin{table}[t]
\centering
\caption{Analysis of Different Masking Ratios}
\label{tab:ratios}
\begin{tabular}{cc||llll}
\hline
 &  \multicolumn{5}{c}{Spatial Transformer} \\

\multirow{5}{*}{} & \multicolumn{1}{l}{} & \multicolumn{1}{c}{0.5} & \multicolumn{1}{c}{0.6} & \multicolumn{1}{c}{0.7} & \multicolumn{1}{c}{0.8} \\
\hline
 & 0.5 & 85.94 & 87.85 & 90.35 & 89.01 \\
 Spectral& 0.6 & 88.19 & 87.71 & 90.39 & 82.45 \\
 Transformer& 0.7 & 84.97 & 86.67 & \textbf{91.30} & 89.62 \\
 & 0.8 & 87.51 & 86.57 & 89.84 & 88.63 \\
 \hline
\end{tabular}
\end{table}

\begin{table}[t]
\centering
\caption{Analysis of Different Spatial Patch Sizes}
\label{tab:patch_sizes}
\begin{tabular}{c||cccc}
\hline 
Metric & 3x3 & 5x5 & 7x7 & 9x9 \\
\hline
OA (\%) & 83.53 &  88.90 & \textbf{91.30} & 87.60 \\
AA (\%) & 88.76 &  93.31 & \textbf{94.30} & 92.77 \\
\textit{k} & 0.8118 &  0.8733 & \textbf{0.9006} & 0.8585 \\
\hline 
\end{tabular}
\end{table}

\vspace{0.1cm}
\noindent \textbf{Patch Size:} The optimal patch size is crucial in utilizing the model's best performance. The patch size determines the amount of spatial information and directly affects the sequence length input to the spatial transformer. To evaluate the model's sensitivity to different patch sizes, we conducted experiments using patch sizes of $3\times3$, $5\times5$, $7\times7$, and $9\times9$, as shown in Table \ref{tab:patch_sizes}. The $3\times3$ patch size resulted in the lowest performance due to insufficient information, while the accuracy increased with increasing patch size until reaching a peak at $7\times7$ and then dropping at $9\times9$. Using large patch sizes introduced information from multiple classes, leading to feature contamination and decreased performance. Based on these results, we selected a patch size of $7\times7$ as optimal for our model.

\vspace{0.2cm}
\noindent \textbf{Grouping Neighbouring Spectral Bands:} Considering the redundancy in spectral dimension, in order to select the optimal use of spectral information during patching in the spectral transformer, we conducted an ablation study by grouping each band along with it's neighbouring bands. Table \ref{tab:band_grouping} shows the performance of the model evaluated by grouping different number of neighbouring bands. Sending single bands as inputs performs the best demonstrating that the inherent transformer network architecture handles the redundancy with the multi-head self-attention mechanism.

\begin{table}[b]
\centering
\caption{Analysis of grouping spectral bands in the spectral transformer}
\label{tab:band_grouping}
\begin{tabular}{c||ccccc}
\hline 

Metric & 1 & 3 & 5 & 7  \\
\hline
OA (\%) & \textbf{91.30} & 90.05 & 87.34 & 85.16 \\
AA (\%) & \textbf{94.30} & 93.81 & 93.31 & 90.54 \\
k & \textbf{0.9006} & 0.8864 & 0.8568 & 0.8293  \\
\hline 
\end{tabular}
\end{table}

\vspace{0.1cm}
\noindent \textbf{Computational Efficiency:} In order to compare the computational complexity of the proposed factorized pretraining approach with MAEST \cite{MAEST}, we conducted a study using the same resources to compute the run times, and the average of five epochs was reported. Table \ref{tab:comp-complexity} shows the number of parameters and the pretraining run time per epoch for both \textit{FactoFormer} and MAEST\cite{MAEST} methods. The autoencoder architecture of MAEST\cite{MAEST} requires heavy decoders to reconstruct the whole input, resulting in significantly high run-time and number of parameters. In contrast, we utilize a single linear layer prediction head to predict the original signals at the masked regions, substantially reducing the number of parameters. The sum of parameters in our spatial and spectral encoders is only one-third compared to MAEST\cite{MAEST}. The run times for both spatial and spectral transformers in our approach are approximately 7 seconds, even though the number of parameters is almost fourfold in the spatial Transformer. The reported times are similar due to the variation in the length of the input sequence. Compared to MAEST\cite{MAEST}, the proposed \textit{FactoFormer} reports a twelve times lower run-time, demonstrating that the factorized pretraining approach is computationally more efficient.

\begin{figure}[t]
    \centering
    \includegraphics[scale=0.35]{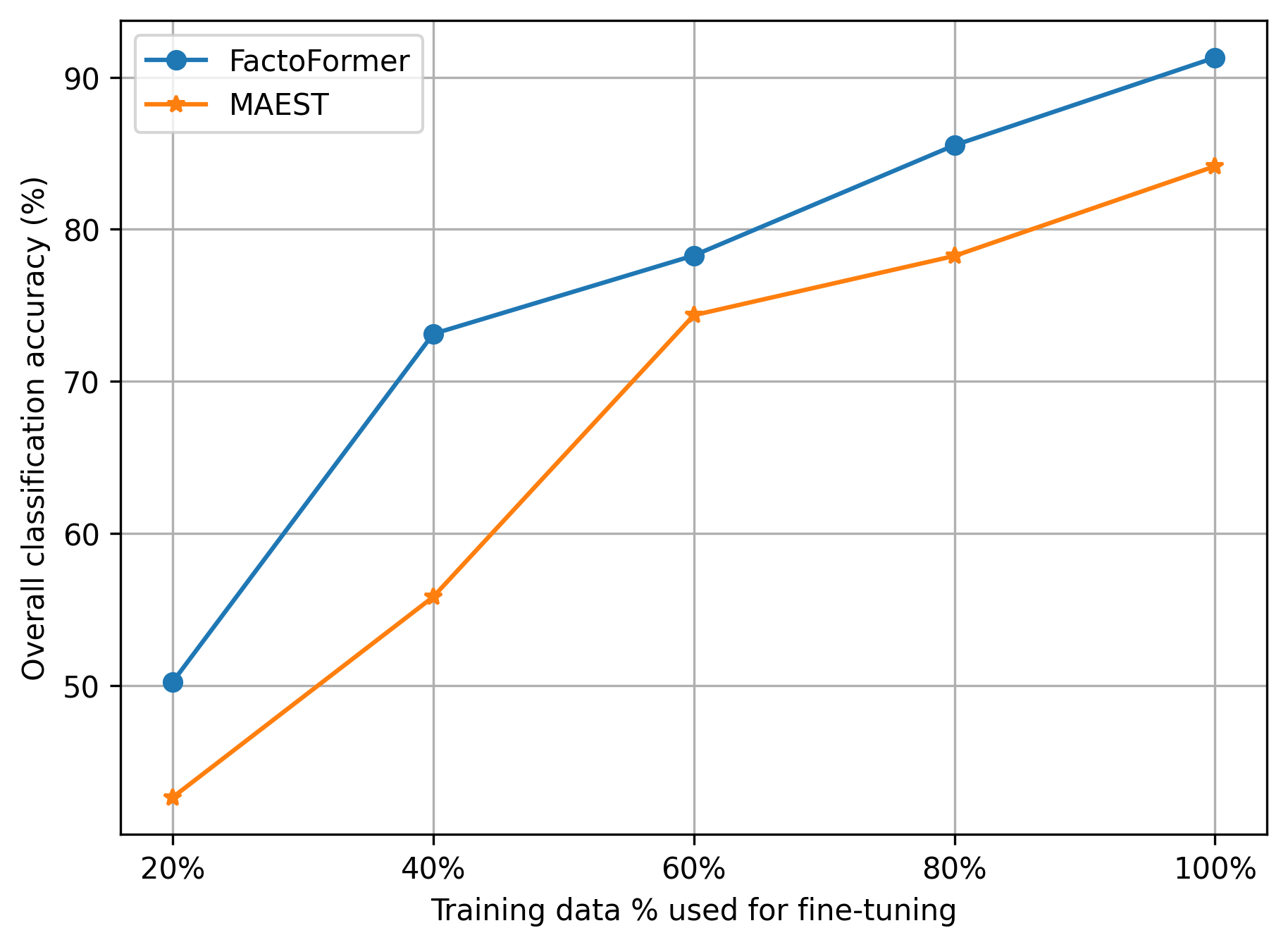}
    \caption{Overall classification accuracy of Indian Pines dataset using different  training data sizes during fine-tuning.} 
    \label{fig:data_percentage}
\end{figure}

\vspace{0.1cm}
\noindent \textbf{Impact of Dataset Size for Fine-Tuning:} We also conducted a study to analyze the performance of the pretrained model by varying the size of training data used during fine-tuning. For this experiment, we used different percentages of data samples from the Indian Pines dataset, ranging from 20\% to 100\%. We compared \textit{FactoFormer} with MAEST \cite{MAEST} and the results are shown in Figure \ref{fig:data_percentage}. Notably, \textit{FactoFormer} consistently outperforms MAEST \cite{MAEST} across all the different data percentages. As we increase the amount of data, the accuracy further improves. This analysis demonstrates that the model learns a better representation through the proposed factorized self-supervised pretraining.

\begin{table}[t]
\centering
\caption{Pretraining Computational Complexity Analysis}
\label{tab:comp-complexity}
\begin{tabular}{l||cc}
\hline
 &  \begin{tabular}[c]{@{}c@{}}Number of\\ Parameters\end{tabular} &  \begin{tabular}[c]{@{}c@{}}Run-time for \\ 1 epoch (s)\end{tabular} \\
 \hline
\textit{FactoFormer} (Spectral Transformer) & 33K & 7.35 \\
\textit{FactoFormer} (Spatial Transformer) & 119K & 7.05 \\
MAEST\cite{MAEST} & 470K & 90.44 \\
\hline
\end{tabular}
\end{table}

%% file: chapters/discussion.tex
\section{Discussion}
\label{sec:Discussion}

\textit{FactoFormer} demonstrates state-of-the-art performance across all six datasets and outperforms other transformer based methods by a significant margin. The main advantage of \textit{FactoFormer} is that it enables factorized self-attention and factorized self-supervised pretraining, resulting in improved performance compared to SpectralFormer \cite{hong2021spectralformer} and MAEST \cite{MAEST}. We highlight the importance of utilizing both spectral and spatial features compared to focusing only on spectra with a more challenging downstream task like precise crop classification using the WHU-Hi dataset. We also show that our pretraining is computationally more efficient than MAEST \cite{MAEST} since we utilize a single linear layer prediction head as the decoder. Additionally, it's worth noting that our models converge in fewer epochs during fine-tuning on all datasets, which further highlights the efficiency of the pretraining approach. Furthermore, our approach consistently outperforms MAEST \cite{MAEST} when fine-tuned with different percentages of training data.

Our pretraining approach shares the same limitation as other available techniques when dealing with data collected from various sensors. Hyperspectral imaging datasets exhibit variations in properties such as wavelength range, spectral resolution, the number of bands, and spatial resolution across various sensors. With our current approach in common with most other reported approaches, we need to pretrain the model every time we use a new dataset collected from a different sensor. For this reason, cross-domain hyperspectral image classification is drawing increasing attention \cite{ref_2}, \cite{ref_3}, \cite{ref_5}. In the case of self-supervised learning, pretraining in the language and vision domains has shown that pretraining on large datasets achieves state-of-the-art performance. Handling sensor variations in HSI and combining datasets for pretraining is an interesting direction for future research.

%% file: chapters/conclusion.tex
\section{Conclusion}
\label{sec:conclusion}

This paper introduced a novel factorized transformer architecture, \textit{FactoFormer}, with self-supervised pretraining for hyperspectral data. The factorized architecture enables factorized self-attention, where attention is computed by individually focusing on spectral and spatial dimensions in each transformer, leading to the learning of salient spatial-spectral representations. The proposed masking strategy within the self-supervised pretraining pipeline has significantly contributed to better representation learning in the proposed model. \textit{FactoFormer} achieves state-of-the-art performance while being computationally more efficient compared to the previous state-of-the-art transformer methods. In future research, we will address the problem of unsupervised domain adaptation by investigating how to transfer the learned representations between different hyperspectral datasets while utilizing common properties, which will enable pretraining on large hyperspectral datasets and fine-tuning on smaller datasets.

%% file: main.bbl
\begin{thebibliography}{10}
\providecommand{\url}[1]{#1}
\csname url@rmstyle\endcsname
\providecommand{\newblock}{\relax}
\providecommand{\bibinfo}[2]{#2}
\providecommand\BIBentrySTDinterwordspacing{\spaceskip=0pt\relax}
\providecommand\BIBentryALTinterwordstretchfactor{4}
\providecommand\BIBentryALTinterwordspacing{\spaceskip=\fontdimen2\font plus
\BIBentryALTinterwordstretchfactor\fontdimen3\font minus \fontdimen4\font\relax}
\providecommand\BIBforeignlanguage[2]{{%
\expandafter\ifx\csname l@#1\endcsname\relax
\typeout{** WARNING: IEEEtran.bst: No hyphenation pattern has been}%
\typeout{** loaded for the language `#1'. Using the pattern for}%
\typeout{** the default language instead.}%
\else
\language=\csname l@#1\endcsname
\fi
#2}}

\bibitem{moghadam2017plant}
P.~Moghadam, D.~Ward, E.~Goan, S.~Jayawardena, P.~Sikka, and E.~Hernandez, ``Plant disease detection using hyperspectral imaging,'' in \emph{2017 International Conference On Digital Image Computing-Techniques and Applications (DICTA)}.\hskip 1em plus 0.5em minus 0.4em\relax IEEE, 2017, pp. 384--391.

\bibitem{GUO2023120}
Y.~Guo, K.~Mokany, C.~Ong, P.~Moghadam, S.~Ferrier, and S.~R. Levick, ``Plant species richness prediction from desis hyperspectral data: A comparison study on feature extraction procedures and regression models,'' \emph{ISPRS Journal of Photogrammetry and Remote Sensing}, vol. 196, pp. 120--133, 2023.

\bibitem{military}
M.~Shimoni, R.~Haelterman, and C.~Perneel, ``Hypersectral imaging for military and security applications: Combining myriad processing and sensing techniques,'' \emph{IEEE Geoscience and Remote Sensing Magazine}, vol.~7, no.~2, pp. 101--117, 2019.

\bibitem{urban_planning}
C.~Weber, R.~Aguejdad, X.~Briottet, J.~Avala, S.~Fabre, J.~Demuynck, E.~Zenou, Y.~Deville, M.~S. Karoui, F.~Z. Benhalouche, \emph{et~al.}, ``Hyperspectral imagery for environmental urban planning,'' in \emph{IGARSS 2018-2018 IEEE International Geoscience and Remote Sensing Symposium}.\hskip 1em plus 0.5em minus 0.4em\relax IEEE, 2018, pp. 1628--1631.

\bibitem{geology}
S.~Peyghambari and Y.~Zhang, ``Hyperspectral remote sensing in lithological mapping, mineral exploration, and environmental geology: an updated review,'' \emph{Journal of Applied Remote Sensing}, vol.~15, no.~3, p. 031501, 2021.

\bibitem{food-quality}
Y.~Liu, H.~Pu, and D.-W. Sun, ``Hyperspectral imaging technique for evaluating food quality and safety during various processes: A review of recent applications,'' \emph{Trends in food science \& technology}, vol.~69, pp. 25--35, 2017.

\bibitem{medical}
U.~Khan, S.~Paheding, C.~P. Elkin, and V.~K. Devabhaktuni, ``Trends in deep learning for medical hyperspectral image analysis,'' \emph{IEEE Access}, vol.~9, pp. 79\,534--79\,548, 2021.

\bibitem{1D-CNN}
W.~Lv and X.~Wang, ``Overview of hyperspectral image classification,'' \emph{Journal of Sensors}, vol. 2020, 2020.

\bibitem{2D-CNN}
X.~Cao, J.~Yao, Z.~Xu, and D.~Meng, ``Hyperspectral image classification with convolutional neural network and active learning,'' \emph{IEEE Transactions on Geoscience and Remote Sensing}, vol.~58, no.~7, pp. 4604--4616, 2020.

\bibitem{chen2016deep}
Y.~Chen, H.~Jiang, C.~Li, X.~Jia, and P.~Ghamisi, ``Deep feature extraction and classification of hyperspectral images based on convolutional neural networks,'' \emph{IEEE Transactions on Geoscience and Remote Sensing}, vol.~54, no.~10, pp. 6232--6251, 2016.

\bibitem{roy2019hybridsn}
S.~K. Roy, G.~Krishna, S.~R. Dubey, and B.~B. Chaudhuri, ``Hybridsn: Exploring 3-d--2-d cnn feature hierarchy for hyperspectral image classification,'' \emph{IEEE Geoscience and Remote Sensing Letters}, vol.~17, no.~2, pp. 277--281, 2019.

\bibitem{mahendren2021reduction}
S.~Mahendren, T.~Fernando, S.~Sridharan, P.~Moghadam, and C.~Fookes, ``Reduction of feature contamination for hyper spectral image classification,'' in \emph{2021 International Conference on Digital Image Computing: Techniques and Applications (DICTA)}, 2021.

\bibitem{hong2021spectralformer}
D.~Hong, Z.~Han, J.~Yao, L.~Gao, B.~Zhang, A.~Plaza, and J.~Chanussot, ``Spectralformer: Rethinking hyperspectral image classification with transformers,'' \emph{IEEE Transactions on Geoscience and Remote Sensing}, vol.~60, pp. 1--15, 2021.

\bibitem{MAEST}
D.~Iba{\~n}ez, R.~Fernandez-Beltran, F.~Pla, and N.~Yokoya, ``Masked auto-encoding spectral-spatial transformer for hyperspectral image classification,'' \emph{IEEE Transactions on Geoscience and Remote Sensing}, 2022.

\bibitem{SSFT}
L.~Sun, G.~Zhao, Y.~Zheng, and Z.~Wu, ``Spectral–spatial feature tokenization transformer for hyperspectral image classification,'' \emph{IEEE Transactions on Geoscience and Remote Sensing}, vol.~60, pp. 1--14, 2022.

\bibitem{MAE}
K.~He, X.~Chen, S.~Xie, Y.~Li, P.~Doll{\'a}r, and R.~Girshick, ``Masked autoencoders are scalable vision learners,'' in \emph{Proceedings of the IEEE/CVF Conference on Computer Vision and Pattern Recognition}, 2022, pp. 16\,000--16\,009.

\bibitem{SimMIM}
Z.~Xie, Z.~Zhang, Y.~Cao, Y.~Lin, J.~Bao, Z.~Yao, Q.~Dai, and H.~Hu, ``Simmim: A simple framework for masked image modeling,'' in \emph{Proceedings of the IEEE/CVF Conference on Computer Vision and Pattern Recognition}, 2022, pp. 9653--9663.

\bibitem{AE-chen2014deep}
Y.~Chen, Z.~Lin, X.~Zhao, G.~Wang, and Y.~Gu, ``Deep learning-based classification of hyperspectral data,'' \emph{IEEE Journal of Selected topics in applied earth observations and remote sensing}, vol.~7, no.~6, pp. 2094--2107, 2014.

\bibitem{mou2017deep}
L.~Mou, P.~Ghamisi, and X.~X. Zhu, ``Deep recurrent neural networks for hyperspectral image classification,'' \emph{IEEE Transactions on Geoscience and Remote Sensing}, vol.~55, no.~7, pp. 3639--3655, 2017.

\bibitem{zhou2019hyperspectral}
F.~Zhou, R.~Hang, Q.~Liu, and X.~Yuan, ``Hyperspectral image classification using spectral-spatial lstms,'' \emph{Neurocomputing}, vol. 328, pp. 39--47, 2019.

\bibitem{DINO}
M.~Caron, H.~Touvron, I.~Misra, H.~J{\'e}gou, J.~Mairal, P.~Bojanowski, and A.~Joulin, ``Emerging properties in self-supervised vision transformers,'' in \emph{Proceedings of the IEEE/CVF International Conference on Computer Vision}, 2021, pp. 9650--9660.

\bibitem{ref_2}
Y.~Zhang, W.~Li, W.~Sun, R.~Tao, and Q.~Du, ``Single-source domain expansion network for cross-scene hyperspectral image classification,'' \emph{IEEE Transactions on Image Processing}, vol.~32, pp. 1498--1512, 2023.

\bibitem{ref_3}
Y.~Zhang, W.~Li, M.~Zhang, Y.~Qu, R.~Tao, and H.~Qi, ``Topological structure and semantic information transfer network for cross-scene hyperspectral image classification,'' \emph{IEEE Transactions on Neural Networks and Learning Systems}, 2021.

\bibitem{ref_5}
Y.~Zhang, W.~Li, M.~Zhang, S.~Wang, R.~Tao, and Q.~Du, ``Graph information aggregation cross-domain few-shot learning for hyperspectral image classification,'' \emph{IEEE Transactions on Neural Networks and Learning Systems}, 2022.

\bibitem{yang2017learning}
J.~Yang, Y.-Q. Zhao, and J.~C.-W. Chan, ``Learning and transferring deep joint spectral--spatial features for hyperspectral classification,'' \emph{IEEE Transactions on Geoscience and Remote Sensing}, vol.~55, no.~8, pp. 4729--4742, 2017.

\bibitem{9103247}
M.~Zhu, L.~Jiao, F.~Liu, S.~Yang, and J.~Wang, ``Residual spectral–spatial attention network for hyperspectral image classification,'' \emph{IEEE Transactions on Geoscience and Remote Sensing}, vol.~59, no.~1, pp. 449--462, 2021.

\bibitem{vit_original}
A.~Dosovitskiy, L.~Beyer, A.~Kolesnikov, D.~Weissenborn, X.~Zhai, T.~Unterthiner, M.~Dehghani, M.~Minderer, G.~Heigold, S.~Gelly, J.~Uszkoreit, and N.~Houlsby, ``An image is worth 16x16 words: Transformers for image recognition at scale,'' \emph{International Conference on Learning Representations}, 2021.

\bibitem{GAHT}
S.~Mei, C.~Song, M.~Ma, and F.~Xu, ``Hyperspectral image classification using group-aware hierarchical transformer,'' \emph{IEEE Transactions on Geoscience and Remote Sensing}, vol.~60, pp. 1--14, 2022.

\bibitem{CASST}
Y.~Peng, Y.~Zhang, B.~Tu, Q.~Li, and W.~Li, ``Spatial--spectral transformer with cross-attention for hyperspectral image classification,'' \emph{IEEE Transactions on Geoscience and Remote Sensing}, vol.~60, pp. 1--15, 2022.

\bibitem{BYOL}
J.-B. Grill, F.~Strub, F.~Altch{\'e}, C.~Tallec, P.~Richemond, E.~Buchatskaya, C.~Doersch, B.~Avila~Pires, Z.~Guo, M.~Gheshlaghi~Azar, \emph{et~al.}, ``Bootstrap your own latent-a new approach to self-supervised learning,'' \emph{Advances in neural information processing systems}, vol.~33, pp. 21\,271--21\,284, 2020.

\bibitem{LiuSSL}
X.~Liu, F.~Zhang, Z.~Hou, L.~Mian, Z.~Wang, J.~Zhang, and J.~Tang, ``Self-supervised learning: Generative or contrastive,'' \emph{IEEE Transactions on Knowledge and Data Engineering}, vol.~35, no.~1, pp. 857--876, 2021.

\bibitem{SimCLR}
T.~Chen, S.~Kornblith, M.~Norouzi, and G.~Hinton, ``A simple framework for contrastive learning of visual representations,'' in \emph{International conference on machine learning}.\hskip 1em plus 0.5em minus 0.4em\relax PMLR, 2020, pp. 1597--1607.

\bibitem{vaswani2017attention}
A.~Vaswani, N.~Shazeer, N.~Parmar, J.~Uszkoreit, L.~Jones, A.~N. Gomez, {\L}.~Kaiser, and I.~Polosukhin, ``Attention is all you need,'' \emph{Advances in neural information processing systems}, vol.~30, 2017.

\bibitem{IP}
M.~F. Baumgardner, L.~L. Biehl, and D.~A. Landgrebe, ``220 band aviris hyperspectral image data set: June 12, 1992 indian pine test site 3,'' \emph{Purdue University Research Repository}, vol.~10, no.~7, p. 991, 2015.

\bibitem{UP}
X.~Huang and L.~Zhang, ``A comparative study of spatial approaches for urban mapping using hyperspectral rosis images over pavia city, northern italy,'' \emph{International Journal of Remote Sensing}, vol.~30, no.~12, pp. 3205--3221, 2009.

\bibitem{Houston2013}
C.~Debes, A.~Merentitis, R.~Heremans, J.~Hahn, N.~Frangiadakis, T.~van Kasteren, W.~Liao, R.~Bellens, A.~Pižurica, S.~Gautama, W.~Philips, S.~Prasad, Q.~Du, and F.~Pacifici, ``Hyperspectral and lidar data fusion: Outcome of the 2013 grss data fusion contest,'' \emph{IEEE Journal of Selected Topics in Applied Earth Observations and Remote Sensing}, vol.~7, no.~6, pp. 2405--2418, 2014.

\bibitem{wuhan_dataset}
Y.~Zhong, X.~Hu, C.~Luo, X.~Wang, J.~Zhao, and L.~Zhang, ``Whu-hi: Uav-borne hyperspectral with high spatial resolution (h2) benchmark datasets and classifier for precise crop identification based on deep convolutional neural network with crf,'' \emph{Remote Sensing of Environment}, vol. 250, p. 112012, 2020.

\bibitem{knn-song2016hyperspectral}
W.~Song, S.~Li, X.~Kang, and K.~Huang, ``Hyperspectral image classification based on knn sparse representation,'' in \emph{2016 IEEE international geoscience and remote sensing symposium (IGARSS)}.\hskip 1em plus 0.5em minus 0.4em\relax IEEE, 2016, pp. 2411--2414.

\bibitem{SVM}
G.~Mercier and M.~Lennon, ``Support vector machines for hyperspectral image classification with spectral-based kernels,'' in \emph{IGARSS 2003.2003 IEEE International Geoscience and Remote Sensing Symposium}, vol.~1.\hskip 1em plus 0.5em minus 0.4em\relax IEEE, 2003, pp. 288--290.

\bibitem{RF}
S.~Amini, S.~Homayouni, and A.~Safari, ``Semi-supervised classification of hyperspectral image using random forest algorithm,'' in \emph{2014 IEEE Geoscience and Remote Sensing Symposium}, 2014, pp. 2866--2869.

\bibitem{miniGCN}
D.~Hong, L.~Gao, J.~Yao, B.~Zhang, A.~Plaza, and J.~Chanussot, ``Graph convolutional networks for hyperspectral image classification,'' \emph{IEEE Transactions on Geoscience and Remote Sensing}, vol.~59, no.~7, pp. 5966--5978, 2020.

\end{thebibliography}
